%% file: main.tex
\documentclass[10pt,twocolumn,letterpaper]{article}

\usepackage{wacv}              
\usepackage{algorithm}
\usepackage{algpseudocode}
\usepackage{pifont}
\usepackage[accsupp]{axessibility}  

\input{preamble}

\definecolor{wacvblue}{rgb}{0.21,0.49,0.74}
\usepackage[pagebackref,breaklinks,colorlinks,allcolors=wacvblue]{hyperref}

\def\wacvPaperID{2129} 
\def\confName{WACV}
\def\confYear{2026}

\title{Hestia: Voxel-Face-Aware Hierarchical Next-Best-View Acquisition for Efficient 3D Reconstruction}

\author{
Cheng-You Lu$^{1}$ \qquad
Zhuoli Zhuang$^{1}$ \qquad
Nguyen Thanh Trung Le$^{1}$ \qquad
Da Xiao$^{1}$ \\
Yu-Cheng Chang$^{1}$ \qquad
Thomas Do$^{1}$ \qquad
Srinath Sridhar$^{2}$ \qquad
Chin-Teng Lin$^{1}$ \\
$^{1}$University of Technology Sydney \qquad
$^{2}$Brown University\\
}

\begin{document}
\maketitle
\input{sec/0_abstract}    
\input{sec/1_intro}

\input{sec/2_related}    
\input{sec/3_method}
\input{sec/4_result}

\input{sec/5_conclusion}
{
    \small
    \bibliographystyle{ieeenat_fullname}
    \bibliography{main}
}
\clearpage
\input{sec/supp}

\end{document}

%% file: preamble.tex
\newcommand{\NA}{--} 
\usepackage{multirow}
\usepackage{array,xcolor,colortbl}
\usepackage{bbm}

\definecolor{lightgreen}{RGB}{198,239,206}
\definecolor{lightyellow}{RGB}{255,235,156}
\definecolor{lightorange}{RGB}{255,229,204}

%% file: sec/0_abstract.tex
\begin{abstract}

Advances in 3D reconstruction and novel view synthesis have enabled efficient and photorealistic rendering.  
However, images for reconstruction are still either largely manual or constrained by simple preplanned trajectories.
To address this issue, recent works propose generalizable next-best-view planners that do not require online learning.
Nevertheless, robustness and performance remain limited across various shapes. 
Hence, this study introduces Voxel-Face-Aware \underline{H}ierarchical N\underline{e}xt-Be\underline{s}\underline{t}-V\underline{i}ew \underline{A}cquisition for Efficient 3D Reconstruction (Hestia\footnote{In Greek mythology, Hestia is the goddess of the hearth, symbolizing home, foundation, and structure, representing stability and guidance in complex systems}), which addresses the shortcomings of the reinforcement learning-based generalizable approaches for five-degree-of-freedom viewpoint prediction. 
Hestia systematically improves the planners through four components: a more diverse dataset to promote robustness, a hierarchical structure to manage the high-dimensional continuous action search space, a close-greedy strategy to mitigate spurious correlations, and a face-aware design to avoid overlooking geometry.
Experimental results show that Hestia achieves non-marginal improvements, with at least a 4\% gain in coverage ratio, while reducing Chamfer Distance by 50\% and maintaining real-time inference. 
In addition, Hestia outperforms prior methods by at least 12\% in coverage ratio with a 5-image budget and remains robust to object placement variations. 
Finally, we demonstrate that Hestia, as a next-best-view planner, is feasible for the real-world application.
Our project page is \href{https://johnnylu305.github.io/hestia_web}{\texttt{https://johnnylu305.github.io/hestia\_web}}.
\end{abstract}

%% file: sec/1_intro.tex
\section{Introduction}
\label{sec:intro}

Multiview-based 3D scene reconstruction~\cite{schoenberger2016sfm, yao2018mvsnet, xie2019pix2vox, murez2020atlas, wang2021multi, sayed2022simplerecon, xu2023unifying, wang2024dust3r, duisterhof2024mast3r, pan2024glomap} and novel view synthesis~\cite{mildenhall2020nerf, yu2021pixelnerf, yu2022plenoxels, mueller2022instant, zheng2024gpsgaussian, gslrm2024, charatan2024pixelsplat, sabour2024spotlesssplats, flynn2024quark, chen2025mvsplat} have been central topics in computer vision. 
These methods leverage multiview information to reconstruct high-fidelity scenes. 
However, data acquisition remains a bottleneck. 
Most data is collected manually, which is time-consuming and labor-intensive, follows preplanned camera trajectories, or relies on non-active capture systems~\cite{xu2023vr, broxton2020immersive, lin2021deep, yoon2020novel, li2022neural, lu2024diva, chen2024x360}.

To reduce human effort, next-best-view (NBV) planning has been explored for active capture~\cite{monica2018contour, zha1997next, liu2018object, hardouin2020surface, hardouin2020next, lin2022active, pan2022activenerf, lee2022uncertainty, zhan2022activermap, jin2023neu, sunderhauf2023density, ran2023neurar, guedon2023macarons, jiang2023fisherrf}. 
Traditional next-best-view methods rely on heuristic rules that can work well in specific scenarios but often fail to transfer because fixed rules or hyperparameters do not adapt across scenes~\cite{chen2024gennbv}. 
Learning-based next-best-view planners, including online-learning and generalizable methods, improve over preplanned trajectories, which frequently miss occluded regions.
Within learning-based approaches, reinforcement learning-based (RL-based) generalizable methods~\cite{peralta2020next, chen2024gennbv}, which pretrain on a dataset to avoid online learning and directly predict viewpoints as actions, show promising results.
This removes candidate-viewpoint sampling, which may potentially miss the best views and slow viewpoint acquisition. 
An occupancy-grid formulation~\cite{chen2024gennbv} further demonstrates strong coverage, viewpoint flexibility, and generalization. 
Nevertheless, performance remains limited and insufficiently robust across diverse object geometries.

To address the shortcomings, we propose Hestia, a Voxel-Face-Aware \underline{H}ierarchical N\underline{e}xt-Be\underline{s}\underline{t}-V\underline{i}ew \underline{A}cquisition for Efficient 3D Reconstruction. 
Hestia actively collects data in object-centric scenes by predicting five-degree-of-freedom (5-DoF) viewpoints (x, y, z, yaw, pitch) from voxel-face observations.
Specifically, Hestia systematically defines the next-best-view task by proposing core components such as dataset choice, observation and reward design, action space, and learning schemes, forming a foundation for the planner.

\begin{figure}[t]  
  \centering
  \includegraphics[width=\linewidth]{figures/voxel_ray.png}
  \caption{\textbf{A voxel is worth more than a ray.} 
  Unlike the RL-based generalizable method~\cite{chen2024gennbv}, Hestia treats each voxel as a cube by considering its six faces, rather than a point. This reduces the information loss inherent in point approximations, ensuring a more accurate representation of the voxel.
  }
  \label{fig:voxel_ray}
\end{figure}

An idea is \underline{\textit{``A voxel is worth more than a ray''}}. 
We incorporate the visibility of the six faces of each voxel into both the observation and the reward function (see~\cref{fig:voxel_ray,sec:methods}). 
Theoretically, if we sample with a one-ray camera in a scene with \(k\) unit cubes and treat each voxel as a point, then from a coupon collector’s perspective~\cite{boneh1997coupon} approximately \(k^{-1/6}\) of the faces will be missed when sampling stops (see Sec.~\ref{asec:theory}). 
Treating each voxel as a cube enables full face coverage, so Hestia accounts for individual voxel-face visibility and captures data more comprehensively. 
This representation adds little computational overhead and still achieves real-time operation at 25 FPS (see~\cref{tab:avg_bench}).

Hestia further improves learning by refining the observation, action, and learning process. 
For observation, Hestia uses the largest dataset that we processed from Objaverse~\cite{Deitke_2023_CVPR, deitke2024objaverse} for the next-best-view task, exposing the policy to a broad range of surface geometries rather than mostly cubic shapes (see Sec.~S8). 
For action, instead of predicting the full 5-DoF next-best view in one step, Hestia adopts a hierarchical structure. 
The policy first predicts a look-at point as the target of attention, then determines the viewpoint position conditioned on this point. 
For learning, Hestia formulates the task as a close-greedy optimization problem in which, given an occupancy grid, it selects the view that maximizes the current coverage ratio. 
The policy relies only on the previous image, the previous camera pose, and the occupancy grid, rather than a long sequence of images and poses~\cite{peralta2020next, chen2024gennbv}. 
We also use a small reward discount factor \(\gamma\) to prioritize immediate improvements without depending on an oversized terminal reward. 
Similar to a greedy algorithm, this reduces spurious correlations\footnote{Spurious correlations~\cite{interpretation1971spurious, kim2024discovering} refer to certain groups contributing to model errors. In this study, spurious correlations refer to large positive future rewards assigned to suboptimal current next-best-view decisions, leading to ineffective policy learning.} between current actions and large future rewards (see Fig.~S8). 
As a result, Hestia reaches higher coverage with fewer images than prior methods~\cite{guedon2023macarons, jin2023neu, chen2024gennbv, peralta2020next}.
Finally, we demonstrate that Hestia, as a next-best-view planner, is feasible for real-world application using a drone with an RGB camera as a mobile agent and a depth predictor~\cite{duisterhof2024mast3r, wang2024dust3r} to convert RGB images into depth maps. 
The contributions of this work are as follows:
\begin{itemize}
\item A RL-based generalizable next-best-view planner that considers voxels as cubes rather than points to avoid geometry overlooking.
\item A hierarchical structure for handling the high-dimensional continuous action space, a larger and more diverse training set for promoting robustness, and a close-greedy strategy for reducing spurious correlations.
\item Comprehensive evaluations on three datasets show that Hestia achieves non-marginal improvements and is suitable for 3D reconstruction under limited acquisition budgets.
\end{itemize}

%% file: sec/2_related.tex
\section{Related Work}
\label{sec:rel}

The literature review mainly focuses on next-best-view methods that are formulated in 5 DoF or assume a drone as an agent.\\
\textbf{Scene-specific next-best-view planners.} Next-best-view planners have demonstrated promising results in active 3D reconstruction by predicting the optimal viewpoint for data capture based on the current state.
Traditional approaches~\cite{monica2018contour, zha1997next, liu2018object, hardouin2020surface, hardouin2020next, dai2020fast, zhou2021fuel} rely on hand-crafted rules to determine the next-best viewpoint. 
For instance, the method~\cite{zha1997next} selected the next-best viewpoint by maximizing a rating function favoring smooth regions, which may overlook fine-grained object details.
Instead, the methods~\cite{monica2018contour, hardouin2020surface} collected data along boundaries between seen and unseen surfaces to capture finer details, but still require handcrafted parameter tuning for each scene.
Another approach~\cite{liu2018object} scanned segmented objects sequentially using a predefined object database, reducing the need for handcrafted tuning, but its performance degrades in cluttered environments due to inaccurate object matching.
Recent advances in deep learning and increased computational power have given rise to learning-based methods~\cite{lin2022active, pan2022activenerf, lee2022uncertainty, zhan2022activermap, jin2023neu, sunderhauf2023density, chen2024gennbv, peralta2020next, ran2023neurar, zeng2022efficient, tao2022seer, wilson2025pop, li2025activesplat}.
Some studies~\cite{lin2022active, sunderhauf2023density} used NeRF~\cite{mildenhall2020nerf} ensembles to estimate uncertainty via model disagreement for viewpoint selection, resulting in linearly increasing computational overhead.
Other works~\cite{pan2022activenerf, ran2023neurar, zeng2022efficient} avoid the computational overhead by incorporating Bayesian-based NeRF~\cite{shen2021stochastic, martin2021nerf} to estimate uncertainty for viewpoint selection.
Meanwhile, other approaches~\cite{lee2022uncertainty, zhan2022activermap} defined the next-best viewpoint as the viewpoint that maximizes the entropy of the density field along the camera rays.
Although these methods have shown outstanding performance in collecting data, they typically require sampling candidate viewpoints.
In addition, their reliance on online learning makes them less suitable for real-time applications. \\
\textbf{Generalizable next-best-view planners.} 
Unlike the aforementioned online-learning approaches, the generalizable methods~\cite{jin2023neu, chen2024gennbv, peralta2020next} avoid the training process for new scenes, thereby enabling faster next-best-view selection. 
Prediction time is important for real-world tasks where a robot may run out of battery within a few minutes.
Among generalizable methods, prior work~\cite{jin2023neu} proposed a Bayesian-based NeRF that selects the next-best viewpoint to maximize view variance without additional training.
However, it still requires candidate viewpoint sampling, resulting in performance unsuitable for real-time applications.
Instead, another line of generalizable next-best-view approaches~\cite{chen2024gennbv, peralta2020next} utilized reinforcement learning to learn a next-best-view planner to bypass the need for sampling candidate viewpoints.
Prior work~\cite{peralta2020next} proposed learning a 3-DoF next-best-view policy using a series of grayscale images as observations.
Subsequently, prior work~\cite{chen2024gennbv} improved upon this method by incorporating occupancy grids into the observations, which provide explicit geometric information. 
This enhancement enabled the development of a 5-DoF next-best-view planner, achieving an outstanding coverage ratio for unknown scenes.

As shown in Tab.~S4, compared to the scene-specific methods, Hestia does not require candidate viewpoint sampling or inference-time optimization, thereby enabling more flexible viewpoint prediction. 
Compared to the generalizable method~\cite{chen2024gennbv}, Hestia treats voxels as cubes rather than points, enabling more comprehensive capture. 
In addition, Hestia adopts a close-greedy scheme to mitigate spurious correlations, introduces a hierarchical structure to model the action space, and uses a more diverse dataset to maintain robust performance across varying object shapes and positions.
Notably, Hestia’s hierarchical structure addresses the challenge of high-dimensional continuous action search in RL-based generalizable next-best-view planning, which differs from traditional methods~\cite{monica2018contour, hardouin2020surface, dai2020fast, zhou2021fuel}.

%% file: sec/3_method.tex
\section{Methods}
\label{sec:methods}

\subsection{RL Problem Definition}
\label{RL_Def}

\begin{figure*}[t]
   \centering
   \includegraphics[width=\linewidth]{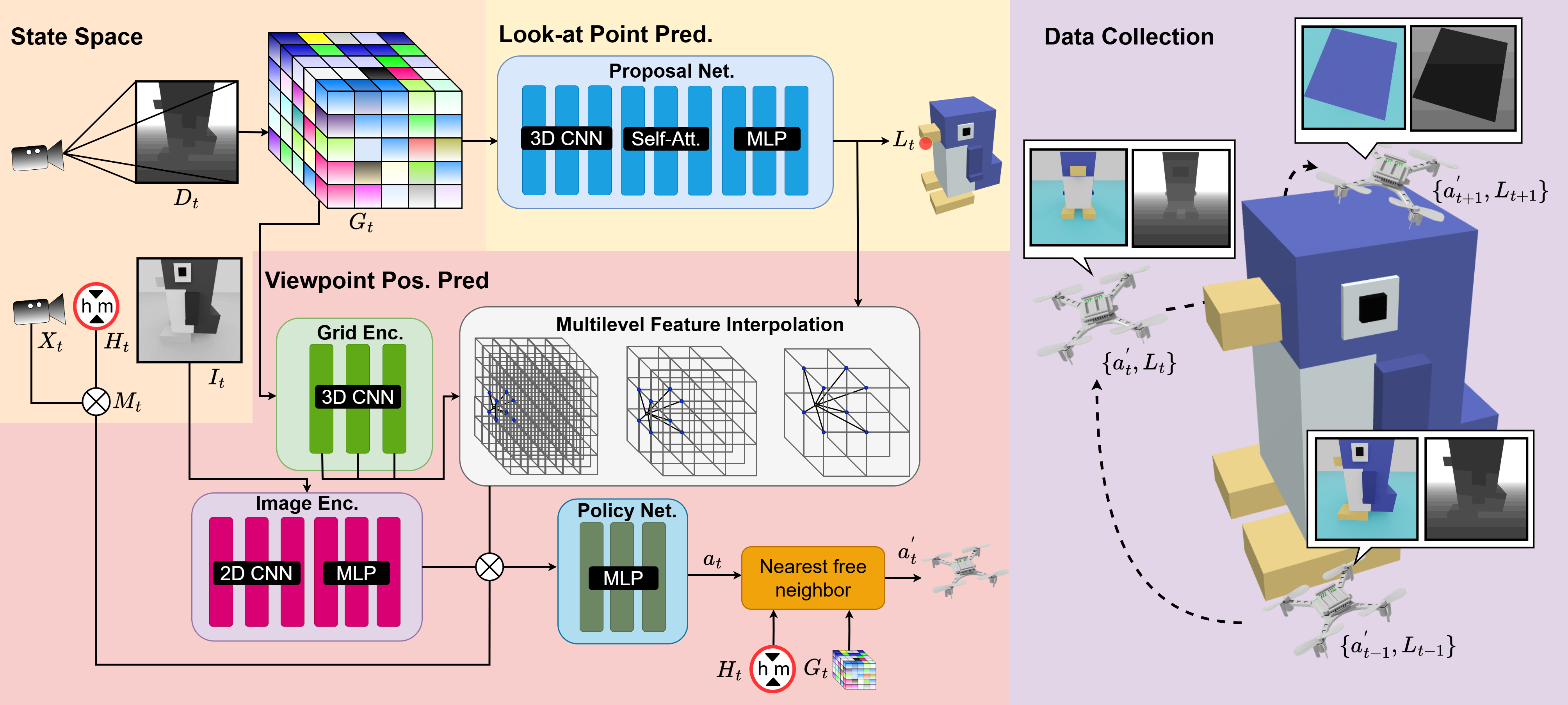}
   \caption{\textbf{Hierarchical structure of Hestia.} Hestia first predicts the camera's look-at point $L_t$ using a proposal neural network that takes grid information $G_t$ processed from the depth image $D_t$ and the camera pose as input. Next, Hestia employs a grid encoder to encode the grid information $G_t$ and performs trilinear interpolation to extract corresponding features from the encoded grid at different layers based on $L_t$. These multilevel interpolated features are then concatenated with the vector information $M_t$ which includes the camera pose $X_t$ and the maximum flyable height, $H_t$ as well as the encoded image features. The image features are extracted using an image encoder, which takes the grayscale image $I_t$ as input. Finally, this combined feature representation is fed into the RL policy model to predict the camera's position $a_t$. Note that Hestia adopts $a'_t$, the nearest collision-free point to $a_t$, as the final camera position to ensure a collision-free viewpoint. Hence, the next-best viewpoint $\{a'_t, L_t\}$ is used for data collection.}
   \label{fig:arch}
\end{figure*}

Our next-best-view task is to identify a 5-DoF viewpoint that maximizes the coverage ratio of an incomplete occupancy grid of the scene. 
The task's goal is similar to greedy methods, which always seek the locally optimal solution.
We formulate the problem as a Markov Decision Process (MDP), denoted by the tuple $\{S, A, P, R, \gamma\}$. 
At each time step $t$, the agent with an RGB-D camera observes a state $s_t$ from the set of all possible states $S$ and chooses an action $a_t$ from the action space $A$. 
The environment then transitions to the subsequent state $s_{t+1}$ according to the probabilities described by $P$, and provides a reward $r_t$.
The magnitude of this reward is determined by the reward function: 
\begin{equation}\label{eq:reward_function}
     R(\cdot \mid s, a): S \times A \to r.
\end{equation}
In reinforcement learning, the main goal is to discover an optimal policy $\pi$ that maximizes the expected sum of discounted rewards, given by: 
\begin{equation}\label{eq:discounted_reward}
     E_t = \sum_{k=0}^{\infty} \gamma^k r_{t+k},
\end{equation}
where $\gamma \in (0, 1)$ is the discount factor.
We set $\gamma$ to 0.1 to align with the greedy-like objective and to avoid spurious correlations from large positive future rewards (see Fig.~S8).
\\
\textbf{State space.} The state space of Hestia is defined as: 
\begin{equation}\label{eq:state_space_hestia}
 S = \Big\{ s_t \;\Big|\; s_t = \big\{ I_t, M_t, G_t, L_t \big\}, \; t \in \mathbb{N} \Big\}
 \end{equation}
where $I_t \in \mathbb{R}^{h \times w}$ is the grayscale image with height $h$ and width $w$, and $L_t \in \mathbb{R}^{3}$ is the camera look-at point. 
The vector $M_t \in \mathbb{R}^{6}$ consists of $X_t \in \mathbb{R}^{5}$, which is the camera position, pitch, and yaw, as well as $H_t \in \mathbb{R}^{1}$, representing the maximum flyable height for the capture. 
Meanwhile, $G_t \in \mathbb{R}^{g \times g \times g \times 10}$ includes the aggregated grid information at resolution $g$, consisting of $O_t \in \mathbb{R}^{g \times g \times g \times 1}$ for the cumulative occupancy grid, $C_t \in \mathbb{R}^{g \times g \times g \times 3}$ for the positional encoding, and $F_t \in \{0, 1\}^{g \times g \times g \times 6}$ for the cumulative face visibility.
The cumulative face visibility is updated iteratively as: 
\begin{equation}
 F_t = f_t \lor F_{t-1}
\end{equation}  
where $F_t$ represents the cumulative face visibility for all voxels up to time $t$, and $f_t \in \{0, 1\}^{g \times g \times g \times 6}$ denotes the current face visibility.
To compute $f_t$, the depth image $D_t$ is unprojected into a voxelized point cloud $V = \{v_i \;|\; i \in \mathbb{N}\}$, where $v_i$ is the $i$-th voxel. 
Each voxel $v_i$ is associated with a viewing direction vector $d_{v_i} \in \mathbb{R}^3$, defined as the vector pointing from the voxel center to the collision-free camera position $a^{'}_{t}$. 
The vector $\mathbf{d}_{v_i}$ is computed as: 
\begin{equation}
\mathbf{d}_{v_i} = \frac{a^{'}_{t} - p_{v_i}}{\|a^{'}_{t} - p_{v_i}\|}
\end{equation}  
where $p_{v_i}$ is the center of voxel $v_i$.
For each voxel $v_i$ and its six outward-facing face normals $n_{i,j} \in \mathbb{R}^3$, the face visibility is determined and aggregated as
\begin{align}
f_t(v_i, j) &= \mathbbm{1}\!\left(d_{v_i} \cdot n_{i,j} > 0\right), \notag \\
&\quad \forall v_i \in V, \; j \in \{1, \dots, 6\}
\end{align}
where $\mathbbm{1}(\cdot)$ is the indicator function.
By iterating over all voxels and their respective faces, $f_t$ is constructed, and the cumulative visibility $F_t$ is updated accordingly.
Although this method cannot handle all face visibilities, the approximation enables efficient computation of face visibility.
Moreover, non-visible voxel faces simply contribute no reward and therefore do not affect the next-best-view selection.
Unlike prior works~\cite{chen2024gennbv}, which consider only $O_t$ and $C_t$ and thereby treat voxels as points, we treat voxels as cubes to mitigate the information loss caused by approximating voxels as points (see~\cref{fig:voxel_ray}).
For details regarding $O_t$ and $C_t$, please refer to the work~\cite{chen2024gennbv}.
\\
\textbf{Action space.} The action space: 
\begin{equation}
A = \Bigl\{ a_t \;\Big|\; a_t \in [-1, 1]^3, \; t \in \mathbb{N} \Bigr\}
\end{equation}
represents the set of possible 3-DoF viewpoints (e.g., camera positions) at each time step $t$, where each coordinate is initially bounded within $[-1, 1]$. 
These coordinates are subsequently normalized to the environment’s scale to ensure appropriate positioning within the scene.
Additionally, the camera’s pitch and yaw are derived from the look-at point and the collision-free action $a^{'}_{t}$ converted from $a_t$ (see~\cref{sec:nbv_net}). 
\\
\textbf{Reward. }
The reward function is defined as:  
\begin{equation}
r_t = R(s_t, a_t) = r_{\text{coverage}}(s_t, a_t) + r_{\text{constraint}}(s_t, a_t)
\end{equation}
where $r_{\text{coverage}}(s_t, a_t)$ encourages the observation of new voxel faces and is expressed as:  
\begin{align}
r_{\text{coverage}}(s_t, a_t) 
&= \frac{\sum_{i=1}^N \sum_{j=1}^6 \left( F_t^{i,j} - F_{t-1}^{i,j} \right) \cdot M_{\text{col}}}{N \cdot 6} \notag \\
&\quad \cdot 0.3
\end{align}
where $F_t^{i,j}$ and $F_{t-1}^{i,j}$ represent the visibility status of the $j$-th face of the $i$-th voxel at time $t$ and $t-1$, respectively. Here, $M_{\text{col}} \in \{0, 1\}$ is a collision indicator, set to $0$ in the event of a collision, thereby preventing any positive reward for invalid actions.
The term $r_{\text{constraint}}(s_t, a_t) = -0.01$ is applied when unsafe or invalid actions occur (see~\cref{asec:rew} for details).
Our reward is based on the face coverage ratio rather than the point coverage ratio to ensure more comprehensive capture (see~\cref{fig:voxel_ray}).  
Furthermore, to prevent spurious correlations, the reward design aligns with a greedy-like objective, which differs significantly from prior works~\cite{chen2024gennbv, peralta2020next} that provide a large goal reward when the coverage ratio reaches a predefined target.

\subsection{Next-Best-View Hierarchical Network}
\label{sec:nbv_net}

The goal of the task is to predict a 5-DoF viewpoint for data collection.  
Directly modeling the 5-DoF viewpoint in the RL continuous action space is challenging due to the high-dimensional search space.  
To address this, Hestia introduces a hierarchical structure to simplify the problem. 
\\
\textbf{Look-at point prediction.} Hestia first predicts the look-at point using a proposal network (see~\cref{fig:arch}), which takes grid information $G_t$ as input to determine \textit{where to look}.  
The proposal network is a 3D convolutional neural network with a self-attention layer to expand the receptive field.
The output is then passed through linear layers to decode the look-at point $L_t$.  
To model the look-at point as a probability distribution, the reparameterization trick is used, treating it as a sample from a normal distribution.  
\\
\textbf{Viewpoint position prediction. } To predict the remaining 3-DoF viewpoint position (e.g., \textit{where to fly}), the grid information is encoded into a multilevel feature grid using a shallow 3D CNN.  
The look-at point $L_t$ is then used to perform trilinear interpolation on the multilevel features from the grid.  
These interpolated features are concatenated with the image embedding, which is extracted by an image encoder, a shallow CNN that takes the grayscale image $I_t$ as input.  
Additionally, the features are concatenated with vector information $M_t$, which includes the camera pose $X_t$ and the maximum flyable height $H_t$.
The combined features are fed into the RL policy model to predict the action $a_t$.  
While the reward function helps constrain $a_t$ to avoid collisions, an additional constraint is applied to ensure a collision-free viewpoint.  
Specifically, $a_t$ is shifted to its nearest collision-free point $a^{'}_{t}$ determined using $G_t$ and $H_t$.  
This adjusted action $a^{'}_{t}$ serves as the final viewpoint position for data capture.  
Thus, Hestia's next-best-view is represented as $\{a^{'}_{t}, L_{t}\}$. 
\\
\textbf{Training loss functions.} 
The look-at point and viewpoint prediction networks can be jointly trained using the \textit{RL reward} due to their connection. 
However, our previous experiments showed no clear benefit from joint training using the \textit{RL reward}. 
To simplify the design, we detach the gradient flow between the networks and train the look-at point prediction network with supervised learning using ground-truth targets.
The entire architecture of Hestia in~\cref{fig:arch} is trained together without any pretraining on other datasets or tasks.
The ground truth look-at point $L_t^\text{gt}$ is computed as the weighted average position of the ground truth uncaptured surface: 
\begin{equation}
L_t^{\text{gt}} = \frac{\sum_{v_i \in U} w_{v_i} \, p_{v_i}}{\sum_{v_i \in U} w_{v_i}}
\end{equation}
where $U$ represents the set of voxels containing ground truth uncaptured faces, and $w_{v_i}$
is defined as the total number of ground truth uncaptured faces within voxel $v_i$:
\begin{equation}
w_{v_i} = \sum_{f \in F_{v_i}^{\text{gt}}} 1
\end{equation}
where $F_{v_i}^{\text{gt}}$ is the set of ground truth uncaptured faces associated with voxel $v_i$.  
Thus, the loss function for the proposal network is formulated as: 
\begin{equation}
\mathcal{L}_{\text{proposal}} = \|L_t -L_t^{\text{gt}}\|^2
\end{equation}
The loss of the viewpoint prediction network is the same as the regular RL loss $\mathcal{L}_{\text{RL}}$ which depends on the RL method used (see~\cref{asec:train_det}), combined with an auxiliary loss: 
\begin{equation}
\mathcal{L}_{\text{aux}} = \|a_t - a^{'}_t\|^2
\end{equation}
to encourage the predicted action $a_t$ to align with the collision-free action $a^{'}_t$.  
Hence, the overall loss function for Hestia is 
\begin{equation}
\mathcal{L}_\text{all} = \mathcal{L}_{\text{RL}} + 0.5 \cdot \mathcal{L}_{\text{aux}} + \mathcal{L}_{\text{proposal}}
\end{equation}

%% file: sec/4_result.tex
\section{Experiments}
\label{sec:exp}

\subsection{Experimental Setup}

Hestia is trained on our processed Objaverse~\cite{deitke2024objaverse, Deitke_2023_CVPR} split to showcase its full capability and on our Houses3K~\cite{peralta2020next} split denoted as Hestia-H3K for fair comparison. 
We use NVIDIA IsaacLab~\cite{mittal2023orbit} to randomly simulate 256 scenes in parallel, with each object scaled up to 8 meters and placed in a 20$\times$20$\times$20m scene.  
Objects are placed at the origin and the four corners for benchmarking.
An RGB-D camera is ahead of the Crazyflie drone, which starts from a random collision-free position oriented toward the object center.
See~\cref{asec:datasets,asec:train_det} for more details.

\subsection{Overall Performance}

This section addresses three questions: \underline{Q1}: Is Hestia’s improvement marginal? \underline{Q2}: Does the method outperform prior works~\cite{jin2023neu, guedon2023macarons, peralta2020next, chen2024gennbv} with and without large-scale training~\cite{deitke2024objaverse, Deitke_2023_CVPR}? \underline{Q3}: Does large-scale training further improve performance? 
To answer these questions, we benchmark on three datasets~\cite{deitke2024objaverse, Deitke_2023_CVPR, wu2023omniobject3d, peralta2020next} comprising 400 diverse shapes, ensuring a comprehensive and fair comparison across methods for the point cloud reconstruction task.
Given the large-scale test set, we select three generalizable baselines~\cite{jin2023neu, chen2024gennbv, peralta2020next} that do not require test-time optimization, along with one online-learning approach~\cite{guedon2023macarons} for benchmarking.
We do not include 3DGS-based online-learning methods~\cite{wilson2025pop, li2025activesplat, khass2025active} as baselines due to differences in data modality, nor methods~\cite{li2025nextbestpath, chen2025gleam} that target non-object-centric scenes with fewer degrees of freedom.

\cref{tab:avg_bench} shows that Hestia not only outperforms prior work on all three datasets, but also achieves at least 4\% and 6\% gains in coverage ratio (CR) and area under the coverage ratio curve (AUC), respectively, while reducing chamfer distance (CD) by 50\% compared to others.
Hence, this answers \underline{Q1}, showing that Hestia’s improvement is not marginal.
Hestia-H3K trained on a smaller, less diverse dataset (Houses3K) still outperforms prior work, demonstrating that the improvement comes not only from large-scale diverse training but also from the proposed designs, thus answering \underline{Q2}.
On both OmniObject3D and Objaverse, Hestia surpasses Hestia-H3K, and even on Hestia-H3K’s own in-distribution set (Houses3K), it achieves slightly better CD, indicating that training on a larger and more diverse dataset provides additional benefits, thus answering \underline{Q3}.

\begin{table*}[thb!]
\centering
\setlength{\tabcolsep}{3.1pt}
\begin{tabular}{cl|
  ccc|
  ccc|
  ccc|
  ccc|c}
\multirow{2}{*}{Train Data} &
\multirow{2}{*}{Method} &
\multicolumn{3}{c|}{OmniObject3D} &
\multicolumn{3}{c|}{Objaverse} &
\multicolumn{3}{c|}{Houses3K} &
\multicolumn{3}{c|}{Overall} &
\multirow{2}{*}{FPS$\uparrow$} \\
& & CR$\uparrow$ & CD$\downarrow$ & AUC$\uparrow$
  & CR$\uparrow$ & CD$\downarrow$ & AUC$\uparrow$
  & CR$\uparrow$ & CD$\downarrow$ & AUC$\uparrow$
  & CR$\uparrow$ & CD$\downarrow$ & AUC$\uparrow$ & \\
    \specialrule{.1em}{.1em}{.1em}
  DTU &  NeU-NBV~\cite{jin2023neu} & 77 & 48 & 73 & 79 & 33 & 75 & 78 & 38 & 75 & 78 & 40 & 74 & $<$1 \\
    \specialrule{.1em}{.1em}{.1em}
    \NA & MACARONS~\cite{guedon2023macarons} & 89 & 17 & 80 & 85 & 24 & 76 & 85 & 26 & 77 & 86 & 22 & 78 & $<$1 \\
    \specialrule{.1em}{.1em}{.1em}
  \multirow{4}{*}{Houses3K} &  ScanRL~\cite{peralta2020next}& 80 & 32 & 75 & 81 & 27 & 78 & 81 & 32 & 76 & 81 & 30 & 76 & 15 \\
    & GenNBV~\cite{chen2024gennbv} & 93 & 12 & \cellcolor{lightyellow}{87} & 91 & 14 & 86 & 92 & 14 & 87 & 92 & 13 & 87 & \cellcolor{lightgreen}{27} \\
    & GenNBV (Rep.)~\cite{chen2024gennbv}& 93 & 11 & \cellcolor{lightyellow}{87} & 92 & 13 & 86 & \cellcolor{lightyellow}{94} & 12 & 89 & 93 & 12 & 87 & \cellcolor{lightgreen}{27} \\
    & Hestia-H3K (Ours) & \cellcolor{lightyellow}{96} & \cellcolor{lightyellow}{6} & \cellcolor{lightgreen}{93}
 & \cellcolor{lightyellow}{95} & \cellcolor{lightyellow}{8} & \cellcolor{lightyellow}{91} & \cellcolor{lightgreen}{97} & \cellcolor{lightyellow}{7} & \cellcolor{lightgreen}{96} & \cellcolor{lightyellow}{96} & \cellcolor{lightyellow}{7} & \cellcolor{lightyellow}{92} & \cellcolor{lightyellow}{25} \\
    \specialrule{.1em}{.1em}{.1em}
    Objaverse & Hestia (Ours)& \cellcolor{lightgreen}{97} & \cellcolor{lightgreen}{4} & \cellcolor{lightgreen}{93} & \cellcolor{lightgreen}{96} & \cellcolor{lightgreen}{7} & \cellcolor{lightgreen}{92} & \cellcolor{lightgreen}{97} & \cellcolor{lightgreen}{6} & \cellcolor{lightyellow}{94} & \cellcolor{lightgreen}{97} & \cellcolor{lightgreen}{6} & \cellcolor{lightgreen}{93} & \cellcolor{lightyellow}{25} \\
    \specialrule{.1em}{.1em}{.1em}
  \end{tabular}
\caption{\textbf{Overall performance on OmniObject3D, Objaverse, and Houses3K with 30 images per object.} Results are reported as mean CR (\%), CD (cm), and AUC (\%) over five object center positions. Hestia and Hestia-H3K outperform prior approaches by at least 4\% and 3\% in CR and by 6\% and 5\% in AUC, respectively, while reducing CD by nearly 50\%. Interestingly, Hestia achieves slightly better CD than Hestia-H3K on in-distribution data (Houses3K).}
\label{tab:avg_bench}
\end{table*}

\subsection{Qualitative Comparisons}
\label{sec:qua}

\begin{figure*}[tbh!]
  \centering
  \includegraphics[width=0.82\linewidth]{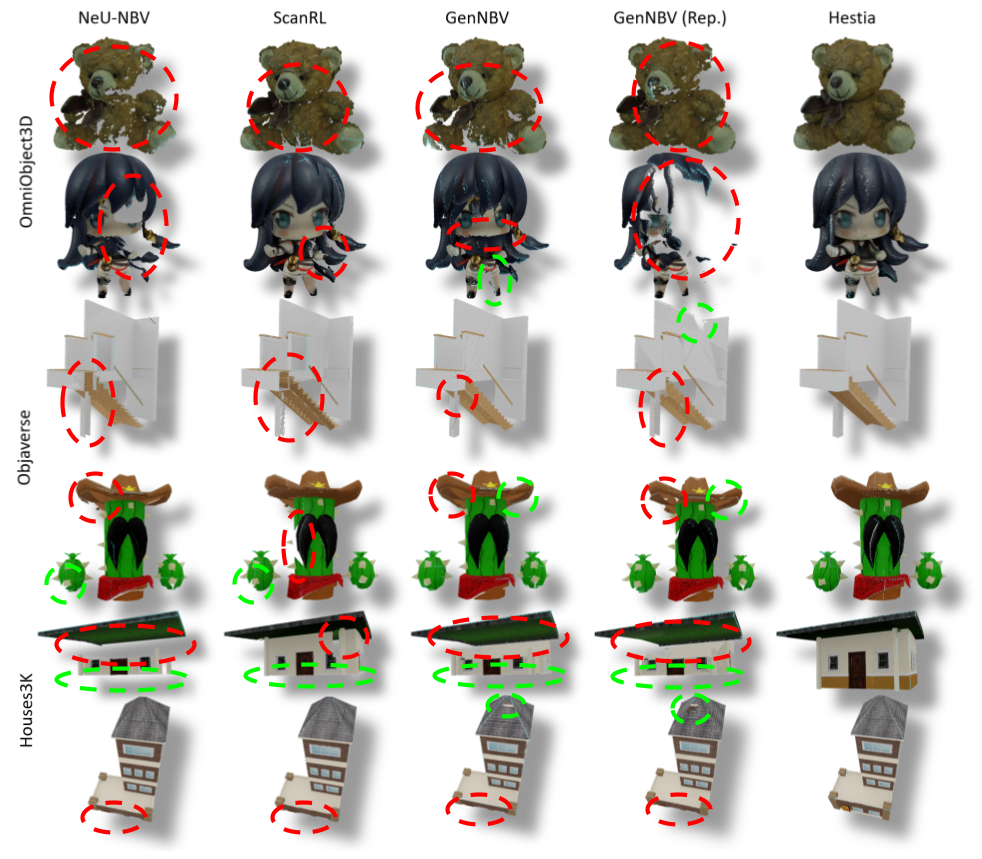}
  \caption{\textbf{Point cloud reconstruction on three datasets.} Hestia’s reconstructions are visibly better than those of prior approaches.}
  \label{fig:vis_all}
\end{figure*}


This section highlights that Hestia’s improvement also extends to visualization in the point cloud reconstruction task. 
\cref{fig:vis_all} shows that Hestia produces more comprehensive point clouds than prior work across diverse object shapes. 
Specifically, prior methods fail to reconstruct parts of the teddy bear and anime figurine from OmniObject3D~\cite{wu2023omniobject3d}, the underside of the stair and the cactus's hat and hands from Objaverse~\cite{deitke2024objaverse, Deitke_2023_CVPR}, and self-occluded structures such as the roof soffit or window from Houses3K~\cite{peralta2020next}.
In addition, Hestia performs well on the complex scenes (see~\cref{fig:vis_all2}).
This improvement is largely attributed to our design, which incorporates a hierarchical structure that better identifies missing parts of objects and models voxels as cubes rather than points, thereby preserving geometric details.
More qualitative results (see~\cref{asec:vis_res,asec:limit}), including failure cases, are provided in the supplementary material, and all reconstruction results are included in the supplementary video. 

\subsection{Translation Robustness}

\begin{figure*}[tbh!]
  \centering
  \includegraphics[width=0.82\linewidth]{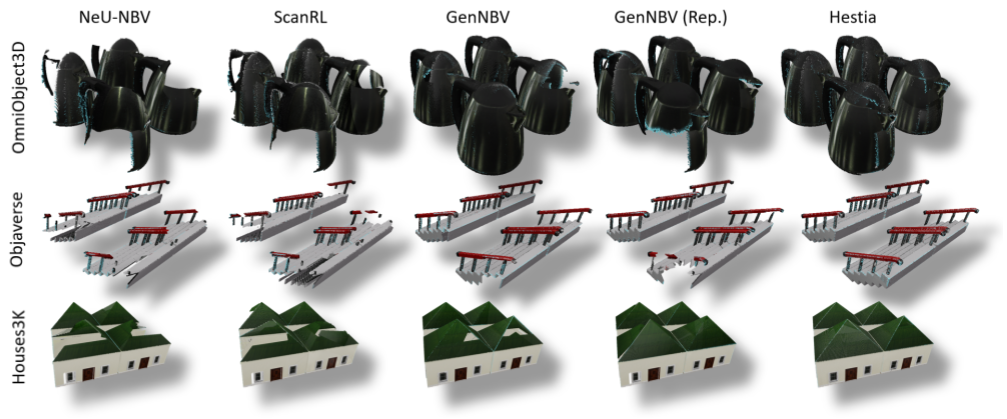}
  \caption{\textbf{Point cloud reconstruction on three datasets.} Hestia’s reconstructions are visibly better than those of prior approaches.}
  \label{fig:vis_shift}
\end{figure*}

This section demonstrates that Hestia maintains robustness when objects are placed at different positions within the scenes. 
\cref{atab:base_bench_omni,atab:base_bench_obj,atab:base_bench_h3k} provide detailed results across different object placement settings.
Hestia exhibits less performance fluctuation, outperforming prior methods on all three datasets.
The qualitative results (see~\cref{fig:vis_shift,asec:novel}) also show that Hestia's reconstruction is more robust.
These results highlight the effectiveness of the hierarchical structure, which first predicts the look-at point and then determines the capture destination.

\subsection{Limited Acquisitions}

This section demonstrates the suitability of Hestia for efficient 3D reconstruction, where only a limited number of views can be acquired. 
\cref{tab:limited_acq} shows that Hestia outperforms prior works by at least 12\% and 5\% in CR under 5-image and 15-image budgets, respectively. 
Specifically, Hestia achieves 92\% CR with only 5 acquisitions, whereas prior work~\cite{chen2024gennbv} requires 15 images to reach comparable performance. 
These gains stem from the close-greedy training strategy, which not only mitigates spurious correlations over time but also enables efficient capture during inference.


\begin{table}[thb!]
\centering
\setlength{\tabcolsep}{3.1pt}
\begin{tabular}{l|cc}
Method & 5 images & 15 images \\
    \specialrule{.1em}{.1em}{.1em}
    NeU-NBV~\cite{jin2023neu} & 71 & 76\\
    ScanRL~\cite{peralta2020next} & 69 & 80\\
    MACARONS~\cite{guedon2023macarons} & 63 & 82\\
    GenNBV~\cite{chen2024gennbv} & 75 & \cellcolor{lightyellow}{91}\\
    GenNBV (Rep.)~\cite{chen2024gennbv} & \cellcolor{lightyellow}{80} & \cellcolor{lightyellow}{91}\\
    \specialrule{.1em}{.1em}{.1em}
    Hestia (Ours) & \cellcolor{lightgreen}{92} & \cellcolor{lightgreen}{96}\\
  \end{tabular}
\caption{\textbf{Mean CR (\%) comparison across the three datasets with limited-view acquisition.} Hestia outperforms prior approaches by at least 12\% and 5\% with a 5-image budget and a 15-image budget. Such efficiency is crucial in real-world power-constrained settings, as robots or agents may exhaust their battery within a short time.}
\label{tab:limited_acq}
\end{table}

\subsection{Inference Speed}

Inference speed is critical for next-best-view planning because robots with onboard cameras must capture images for 3D reconstruction before their batteries are depleted. 
As shown in \cref{tab:avg_bench}, Hestia achieves 25 FPS, which is suitable for real-time deployment. 
Modeling voxels as cubes rather than points does not significantly reduce inference speed. 
The speed also demonstrates the advantage of RL-based generalizable next-best-view approaches, since using a policy model to directly predict viewpoints removes the need to sample candidate views for prediction.

\subsection{Ablation Study}
\label{sec:abl}

This section evaluates the effectiveness of Hestia’s core components through an ablation study (see~\cref{tab:ablation}). 
We investigate three key ideas: face-aware design, a close-greedy training scheme, and a hierarchical structure. 
For the non-hierarchical variant, the encoded grid information is fed directly into the policy model to predict 5-DoF viewpoints without feature interpolation, since interpolation requires the look-at point. 
Applying the close-greedy strategy or the hierarchical structure alone eases the training process, whereas face-aware observation alone may make training more difficult but still provides complementary information. 
Thus, the close-greedy strategy and hierarchical structure yield stronger gains when applied individually, while combining them with face-aware observation further enhances stability and capture quality. 
Overall, each component contributes to performance improvements, and integrating all three delivers the best results.

\begin{table}[thb!]
  \centering
\begin{tabular}{@{}l|ccc|cc|c}
    & Face & Greedy & Hier. & CR $\uparrow$ & CD $\downarrow$ & \#Pa. \\
    \specialrule{.1em}{.1em}{.1em}
    \multirow{9}{*}{Hestia} &  & & & 88 & 20 & 6.2M \\
    \cmidrule{2-7}
    & \checkmark & & & 90 & 17 & 6.2M \\
    & & \checkmark & & 92 & 13  & 6.2M \\
    & & & \checkmark & 94 & 11 & 4.9M \\
    \cmidrule{2-7}
    & \checkmark & \checkmark & & 95 & 9 & 6.2M \\
    & & \checkmark & \checkmark & 95 & 8 & 4.9M \\
    & \checkmark & & \checkmark & 95 & 9 & 4.9M \\
    \cmidrule{2-7}
    & \checkmark & \checkmark & \checkmark & 96 & 7 & 4.9M\\
    \specialrule{.1em}{.1em}{.1em}
  \end{tabular}
\caption{\textbf{Ablations on Objaverse~\cite{deitke2024objaverse, Deitke_2023_CVPR}.} Integrating the proposed ideas yields the best performance with fewer parameters.}
  \label{tab:ablation}
\end{table}

\subsection{Application}

\begin{figure*}[tbh!]
  \centering
  \includegraphics[width=0.82\linewidth]{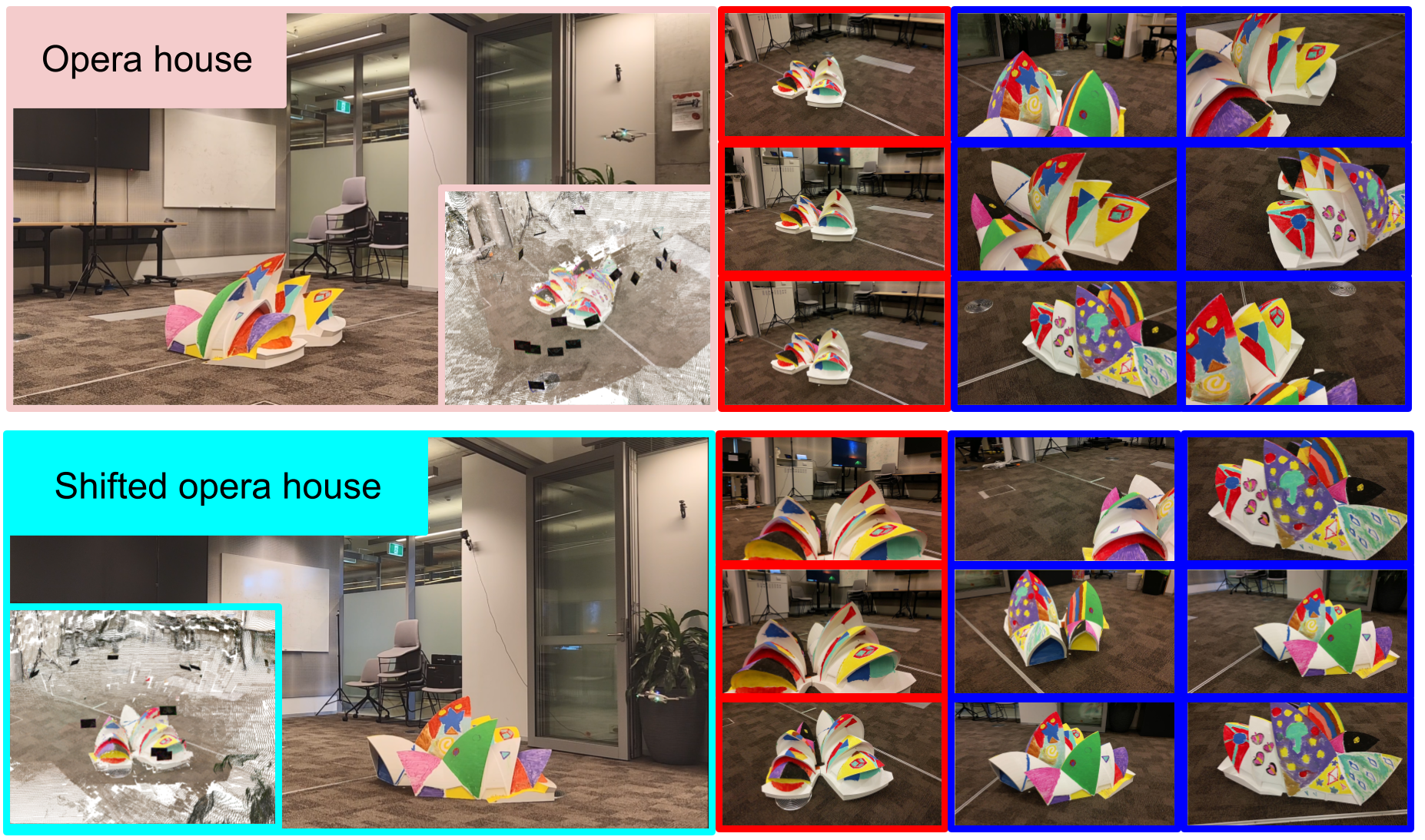}
  \caption{\textbf{Real-world demonstration of non-shifted and shifted scenes.} Red boxes indicate manually initialized viewpoints, while blue boxes denote viewpoints predicted by Hestia. The results demonstrate Hestia’s feasibility in real-world environments.}
  \label{fig:vis_real}
\end{figure*}


This section demonstrates that Hestia is feasible for real-world deployment even without a depth camera. 
We use a drone equipped with an RGB camera as the mobile agent for data collection and employ a depth predictor~\cite{duisterhof2024mast3r, wang2024dust3r} to convert multi-view RGB images into depth maps. 
The first three images are manually selected to synchronize the real-world and virtual-world settings. 
As shown in \cref{fig:vis_real}, Hestia successfully operates in real-world scenarios for both shifted and non-shifted scenes.
It is worth noting that some prior works~\cite{peralta2020next, chen2024gennbv, jin2023neu, guedon2023macarons} report only simulation results, and their real-world feasibility remains unknown. 
Please see~\cref{asec:realworld_res,asec:realworld_sys} for more details.

%% file: sec/5_conclusion.tex
\section{Conclusion}

We present Hestia, voxel-face-aware hierarchical next-best-view acquisition for efficient 3D reconstruction. 
Hestia addresses the high-dimensional action space by separately predicting look-at points and camera positions.
Treating voxels as cubes enables more comprehensive capture, improving coverage ratios. 
The close-greedy design mitigates spurious correlations, ensuring efficient policy learning. 
Trained on a more diverse dataset, Hestia is robust across varied object-centric scenes.
Evaluations on three datasets validate that Hestia’s improvements are not marginal. 
Finally, the integration into a real-world drone system highlights its feasibility. 
As discussed in the limitations and future steps section (\cref{asec:limit}), one important step is extending to multi-agent settings to further improve efficiency.

\section{Acknowledgements}

This work was supported in part by the Australian Research Council (ARC) under discovery grant DP220100803 and DP250103612 and ARC Research Hub for Human-Robot Teaming for Sustainable and Resilient Construction (ITRH) grant IH240100016, and Australian National Health and Medical Research Council (NHMRC) Ideas Grant APP2021183. Research was also sponsored in part by the Australia Advanced Strategic Capabilities Accelerator (ASCA) under Contract No. P18-650825 and ASCA EDT DA ID12994, and the Australian Defence Science Technology Group (DSTG) under Agreement No: 12549.
We thank Yu-Lun Liu for submission guidance, Yi-Shan Hung for proofreading and demo narration, Meredith Porte for additional narration, and Xiao Chen for GenNBV~\cite{chen2024gennbv} discussions.

%% file: sec/supp.tex
\newcommand{\xmark}{\ding{55}} 
\renewcommand{\theequation}{S\arabic{equation}}
\renewcommand{\thefigure}{S\arabic{figure}}
\renewcommand{\thetable}{S\arabic{table}}
\renewcommand{\thesection}{S\arabic{section}}
\renewcommand{\thesubsection}{S\arabic{section}.\arabic{subsection}}
\renewcommand{\algorithmiccomment}[1]{\hspace{0.5em}\(\triangleright\) #1}
\setcounter{figure}{0}
\setcounter{table}{0}
\setcounter{equation}{0}
\setcounter{section}{0}
\setcounter{subsection}{0}
\setcounter{subsubsection}{0}

\section*{Supplementary Materials}

In this supplementary material, we present the theoretical grounding for treating voxels as cubes in~\cref{asec:theory}, additional quantitative results in~\cref{asec:bench}, further qualitative results in~\cref{asec:vis_res}, more real-world demonstration results in~\cref{asec:realworld_res}, details of the reward design in~\cref{asec:rew}, the novelty of the proposed components in~\cref{asec:novel}, the impact of spurious correlations in~\cref{asec:spur}, dataset details and preparation in~\cref{asec:datasets}, the real-world system setup and associated costs in~\cref{asec:realworld_sys}, training and testing details in~\cref{asec:train_det}, and limitations in~\cref{asec:limit}.

\section{Theoretical Grounding}
\label{asec:theory}

Although treating a voxel as a cube rather than a point straightforwardly avoids overlooking surface geometry, we formalize the benefit through the following \textit{constrained example} from a theoretical perspective.
Consider a scene composed of \(k\) unit cubes and a 1–ray camera that emits a single ray per sample, where each ray is assumed to intersect one of the cubes in the scene.
We contrast two sampling rules:
\begin{itemize}
  \item \textbf{Scenario 1 (voxel as a point).} Continue sampling until every cube has been intersected by at least one ray.
  \item \textbf{Scenario 2 (voxel as a cube).} Continue sampling until every face of every cube has been intersected by at least one ray.
\end{itemize}

Our goal is to compare the expected face visibility achieved after the scenarios terminate.
Hitting each of the \(k\) cubes at least once can be treated as a classical coupon–collector problem~\cite{boneh1997coupon}, whose expectation is:
\begin{equation}
  k\Bigl(1+\frac{1}{2}+\dots+\frac{1}{k}\Bigr)
    \approx k\ln k .
  \label{eq:expected_rays_vox}
\end{equation}

Every ray that hits a cube intersects one of its six faces uniformly at random, so each ray can be viewed as a draw from:
\begin{equation}
  N \;=\; 6k
  \label{eq:total_faces}
\end{equation}
distinct faces.
After \(n\) rays, the probability that a specific face is still unseen is:
\begin{equation}
  \Bigl(1-\frac{1}{6k}\Bigr)^{n}
    \;\approx\;
    e^{-\frac{n}{6k}},
  \qquad 6k\gg 1.
  \label{eq:face_unseen_prob}
\end{equation}
Hence, through~\cref{eq:expected_rays_vox,eq:face_unseen_prob}, we know that after Scenario 1 stops, the ratio of the expected non-visible faces is: 
\begin{equation}
    e^{-\frac{ k\ln k}{6k}} = k^{-\frac{1}{6}}.
  \label{eq:face_unseen_ratio}
\end{equation}
If \(k=8000\), roughly 22.3\% of the faces remain unseen for Scenario 1, while Scenario 2 can cover all the faces. 
This theoretical result further motivates treating voxels as cubes rather than points when designing next‑best‑view policies.

\section{Benchmark Details}
\label{asec:bench}

\begin{figure*}[tbh!]
  \centering
  \includegraphics[width=\linewidth]{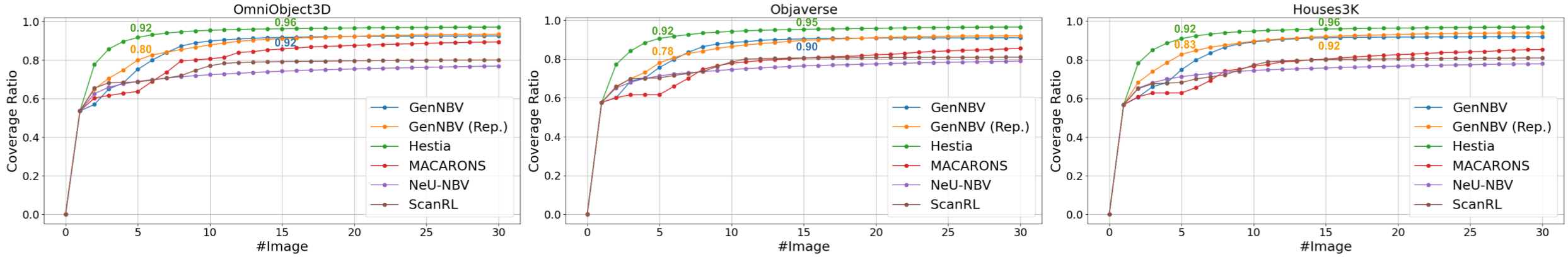}
  \caption{\textbf{CR curves on three datasets.} Hestia outperforms prior approaches by nearly 10\% and 5\% in the first five captures and the first fifteen captures, respectively. The efficiency is particularly significant in real-world power-constrained scenarios, where a robot or agent may run out of battery in a short time.}
  \label{fig:curve}
\end{figure*}

In this section, we present the detailed coverage ratio (CR), Chamfer Distance (CD), and area under the coverage ratio curve (AUC) from~\cref{tab:avg_bench}, broken down by each object position setting in~\cref{atab:base_bench_omni,atab:base_bench_obj,atab:base_bench_h3k}.
Hestia outperforms other baselines across all object position settings in the OmniObject3D~\cite{wu2023omniobject3d}, Objaverse~\cite{Deitke_2023_CVPR}, and Houses3K~\cite{peralta2020next} test splits. 
Moreover, Hestia is the only methods that demonstrate robust performance across all object position settings on all three datasets, with less than a 1\% coverage ratio difference across different object configurations.
For efficient 3D reconstruction,~\cref{fig:curve} shows that Hestia outperforms other methods by nearly 10\% and 5\% in the first five and fifteen captures, respectively. 
This efficiency is especially important in real-world power-constrained scenarios, where a robot or agent may quickly exhaust its battery.
\begin{table*}[thb!]
  \centering
  \setlength{\tabcolsep}{3.1pt}
\begin{tabular}{l|
  ccc|ccc|ccc|ccc|ccc}
     & \multicolumn{15}{c}{OmniObject3D Test} \\
    \specialrule{.1em}{.1em}{.1em}
     \multirow{2}{*}{Method} 
     & \multicolumn{3}{c|}{(0, 0)} 
     & \multicolumn{3}{c|}{(4, 4)} 
     & \multicolumn{3}{c|}{(4, -4)} 
     & \multicolumn{3}{c|}{(-4, 4)} 
     & \multicolumn{3}{c}{(-4, -4)} \\
     & CR$\uparrow$ & CD$\downarrow$ & AUC$\uparrow$
     & CR$\uparrow$ & CD$\downarrow$ & AUC$\uparrow$
     & CR$\uparrow$ & CD$\downarrow$ & AUC$\uparrow$
     & CR$\uparrow$ & CD$\downarrow$ & AUC$\uparrow$
     & CR$\uparrow$ & CD$\downarrow$ & AUC$\uparrow$ \\
    \specialrule{.1em}{.1em}{.1em}
     NeU-NBV~\cite{jin2023neu} & 88 & 24 & 83 & 73 & 51 & 70 & 75 & 75 & 70 & 72 & 53 & 69 & 76 & 38 & 71 \\
    \specialrule{.1em}{.1em}{.1em}
     ScanRL~\cite{peralta2020next} & 87 & 21 & 80 & 79 & 37 & 76 & 84 & 21 & 78 & 72 & 47 & 71 & 76 & 35 & 72 \\
    MACARONS~\cite{guedon2023macarons} & 86 & 19 & 75 & 93 & 13 & 86 & 91 & 14 & 83 & \cellcolor{lightyellow}{94} & \cellcolor{lightyellow}{10} & 88 & 80 & 29 & 69 \\
    GenNBV~\cite{chen2024gennbv} & 92 & 12 & \cellcolor{lightyellow}{88} & 92 & 14 & 86 & 91 & 15 & 83 & \cellcolor{lightyellow}{94} & \cellcolor{lightyellow}{10} & \cellcolor{lightyellow}{89} & \cellcolor{lightyellow}{94} & \cellcolor{lightyellow}{9} & \cellcolor{lightyellow}{88} \\
    GenNBV (Rep.)~\cite{chen2024gennbv} & \cellcolor{lightyellow}{93} & \cellcolor{lightyellow}{10} & 87 & \cellcolor{lightyellow}{94} & \cellcolor{lightyellow}{9} & \cellcolor{lightyellow}{89} & \cellcolor{lightyellow}{92} & \cellcolor{lightyellow}{12} & \cellcolor{lightyellow}{85} & \cellcolor{lightyellow}{94} & \cellcolor{lightyellow}{10} & 88 & 92 & 11 & 85 \\
    \specialrule{.1em}{.1em}{.1em}
    Hestia (Ours) & \cellcolor{lightgreen}{97} & \cellcolor{lightgreen}{4} & \cellcolor{lightgreen}{94} 
           & \cellcolor{lightgreen}{97} & \cellcolor{lightgreen}{4} & \cellcolor{lightgreen}{93} 
           & \cellcolor{lightgreen}{97} & \cellcolor{lightgreen}{5} & \cellcolor{lightgreen}{92} 
           & \cellcolor{lightgreen}{97} & \cellcolor{lightgreen}{4} & \cellcolor{lightgreen}{93} 
           & \cellcolor{lightgreen}{96} & \cellcolor{lightgreen}{5} & \cellcolor{lightgreen}{93} \\
    \specialrule{.1em}{.1em}{.1em}
  \end{tabular}
\caption{\textbf{CR (\%) / CD (cm) / AUC (\%) comparison on the OmniObject3D test set.} Hestia outperforms other methods and is more robust across different object position settings.}
\label{atab:base_bench_omni}
\end{table*}
\begin{table*}[thb!]
  \centering
  \setlength{\tabcolsep}{3.1pt}
\begin{tabular}{l|ccc|ccc|ccc|ccc|ccc}
     & \multicolumn{15}{c}{Objaverse Test} \\
    \specialrule{.1em}{.1em}{.1em}
     \multirow{2}{*}{Method} 
     & \multicolumn{3}{c|}{(0, 0)} 
     & \multicolumn{3}{c|}{(4, 4)} 
     & \multicolumn{3}{c|}{(4, -4)} 
     & \multicolumn{3}{c|}{(-4, 4)} 
     & \multicolumn{3}{c}{(-4, -4)} \\
     & CR$\uparrow$ & CD$\downarrow$ & AUC$\uparrow$
     & CR$\uparrow$ & CD$\downarrow$ & AUC$\uparrow$
     & CR$\uparrow$ & CD$\downarrow$ & AUC$\uparrow$
     & CR$\uparrow$ & CD$\downarrow$ & AUC$\uparrow$
     & CR$\uparrow$ & CD$\downarrow$ & AUC$\uparrow$ \\
    \specialrule{.1em}{.1em}{.1em}
     NeU-NBV~\cite{jin2023neu} & 88 & 20 & 83 & 76 & 41 & 74 & 77 & 31 & 72 & 77 & 41 & 74 & 78 & 31 & 72 \\
    \specialrule{.1em}{.1em}{.1em}
     ScanRL~\cite{peralta2020next} & 87 & 18 & 82 & 80 & 31 & 78 & 82 & 23 & 77 & 77 & 37 & 76 & 80 & 27 & 76 \\
    MACARONS~\cite{guedon2023macarons} & 88 & 17 & 79 & 91 & 16 & 86 & \cellcolor{lightyellow}{91} & \cellcolor{lightyellow}{15} & \cellcolor{lightyellow}{83} & 75 & 42 & 70 & 78 & 30 & 64 \\
    GenNBV~\cite{chen2024gennbv} & \cellcolor{lightyellow}{94} & \cellcolor{lightyellow}{11} & \cellcolor{lightyellow}{89} & 90 & 15 & 85 & 89 & 17 & 82 & \cellcolor{lightyellow}{93} & \cellcolor{lightyellow}{12} & \cellcolor{lightyellow}{89} & \cellcolor{lightyellow}{91} & \cellcolor{lightyellow}{13} & \cellcolor{lightyellow}{86} \\
    GenNBV (Rep.)~\cite{chen2024gennbv} & \cellcolor{lightyellow}{94} & \cellcolor{lightyellow}{11} & 88 & \cellcolor{lightyellow}{92} & \cellcolor{lightyellow}{12} & \cellcolor{lightyellow}{88} & 90 & \cellcolor{lightyellow}{15} & 82 & 92 & \cellcolor{lightyellow}{12} & 88 & 90 & 14 & 84 \\
    \specialrule{.1em}{.1em}{.1em}
    Hestia (Ours) & \cellcolor{lightgreen}{96} & \cellcolor{lightgreen}{7} & \cellcolor{lightgreen}{93} 
           & \cellcolor{lightgreen}{96} & \cellcolor{lightgreen}{7} & \cellcolor{lightgreen}{92} 
           & \cellcolor{lightgreen}{96} & \cellcolor{lightgreen}{6} & \cellcolor{lightgreen}{91} 
           & \cellcolor{lightgreen}{96} & \cellcolor{lightgreen}{7} & \cellcolor{lightgreen}{93} 
           & \cellcolor{lightgreen}{96} & \cellcolor{lightgreen}{8} & \cellcolor{lightgreen}{91} \\
    \specialrule{.1em}{.1em}{.1em}
  \end{tabular}
\caption{\textbf{CR (\%) / CD (cm) / AUC (\%) comparison on the Objaverse test set.} Hestia outperforms other methods and is more robust across different object position settings.}
\label{atab:base_bench_obj}
\end{table*}
\begin{table*}[thb!]
  \centering
  \setlength{\tabcolsep}{3.1pt}
\begin{tabular}{l|
  ccc|ccc|ccc|ccc|ccc}
     & \multicolumn{15}{c}{Houses3K Test} \\
    \specialrule{.1em}{.1em}{.1em}
     \multirow{2}{*}{Method} 
     & \multicolumn{3}{c|}{(0, 0)} 
     & \multicolumn{3}{c|}{(4, 4)} 
     & \multicolumn{3}{c|}{(4, -4)} 
     & \multicolumn{3}{c|}{(-4, 4)} 
     & \multicolumn{3}{c}{(-4, -4)} \\
     & CR$\uparrow$ & CD$\downarrow$ & AUC$\uparrow$
     & CR$\uparrow$ & CD$\downarrow$ & AUC$\uparrow$
     & CR$\uparrow$ & CD$\downarrow$ & AUC$\uparrow$
     & CR$\uparrow$ & CD$\downarrow$ & AUC$\uparrow$
     & CR$\uparrow$ & CD$\downarrow$ & AUC$\uparrow$ \\
    \specialrule{.1em}{.1em}{.1em}
     NeU-NBV~\cite{jin2023neu} & 85 & 26 & 80 & 73 & 52 & 71 & 81 & 30 & 77 & 72 & 53 & 70 & 80 & 32 & 76 \\
    \specialrule{.1em}{.1em}{.1em}
     ScanRL~\cite{peralta2020next} & 86 & 21 & 79 & 77 & 40 & 75 & 88 & 19 & 82 & 72 & 50 & 69 & 81 & 28 & 76 \\
    MACARONS~\cite{guedon2023macarons} & 87 & 20 & 78 & 92 & 16 & 87 & 93 & 14 & 85 & 71 & 51 & 66 & 81 & 30 & 68 \\
    GenNBV~\cite{chen2024gennbv} & \cellcolor{lightyellow}{94} & \cellcolor{lightyellow}{10} & \cellcolor{lightyellow}{89} & 90 & 18 & 83 & 91 & 16 & 84 & \cellcolor{lightyellow}{94} & 12 & \cellcolor{lightyellow}{89} & \cellcolor{lightyellow}{93} & \cellcolor{lightyellow}{12} & \cellcolor{lightyellow}{89} \\
    GenNBV (Rep.)~\cite{chen2024gennbv} & \cellcolor{lightyellow}{94} & 12 & 88 
                                         & \cellcolor{lightyellow}{95} & \cellcolor{lightyellow}{10} & \cellcolor{lightyellow}{90} 
                                         & \cellcolor{lightyellow}{94} & \cellcolor{lightyellow}{13} & \cellcolor{lightyellow}{88} 
                                         & \cellcolor{lightyellow}{94} & \cellcolor{lightyellow}{11} & \cellcolor{lightyellow}{89} 
                                         & 91 & 14 & 88 \\
    \specialrule{.1em}{.1em}{.1em}
    Hestia (Ours) & \cellcolor{lightgreen}{97} & \cellcolor{lightgreen}{5} & \cellcolor{lightgreen}{96} 
           & \cellcolor{lightgreen}{97} & \cellcolor{lightgreen}{5} & \cellcolor{lightgreen}{93} 
           & \cellcolor{lightgreen}{97} & \cellcolor{lightgreen}{7} & \cellcolor{lightgreen}{93} 
           & \cellcolor{lightgreen}{97} & \cellcolor{lightgreen}{5} & \cellcolor{lightgreen}{94} 
           & \cellcolor{lightgreen}{97} & \cellcolor{lightgreen}{7} & \cellcolor{lightgreen}{93} \\
    \specialrule{.1em}{.1em}{.1em}
  \end{tabular}
\caption{\textbf{CR (\%) / CD (cm) / AUC (\%) comparison on the Houses3K test set.} Hestia outperforms other methods and is more robust across different object position settings.}
\label{atab:base_bench_h3k}
\end{table*}

\section{Qualitative Results}
\label{asec:vis_res}
\begin{figure*}[tbh!]
  \centering
  \includegraphics[width=\linewidth]{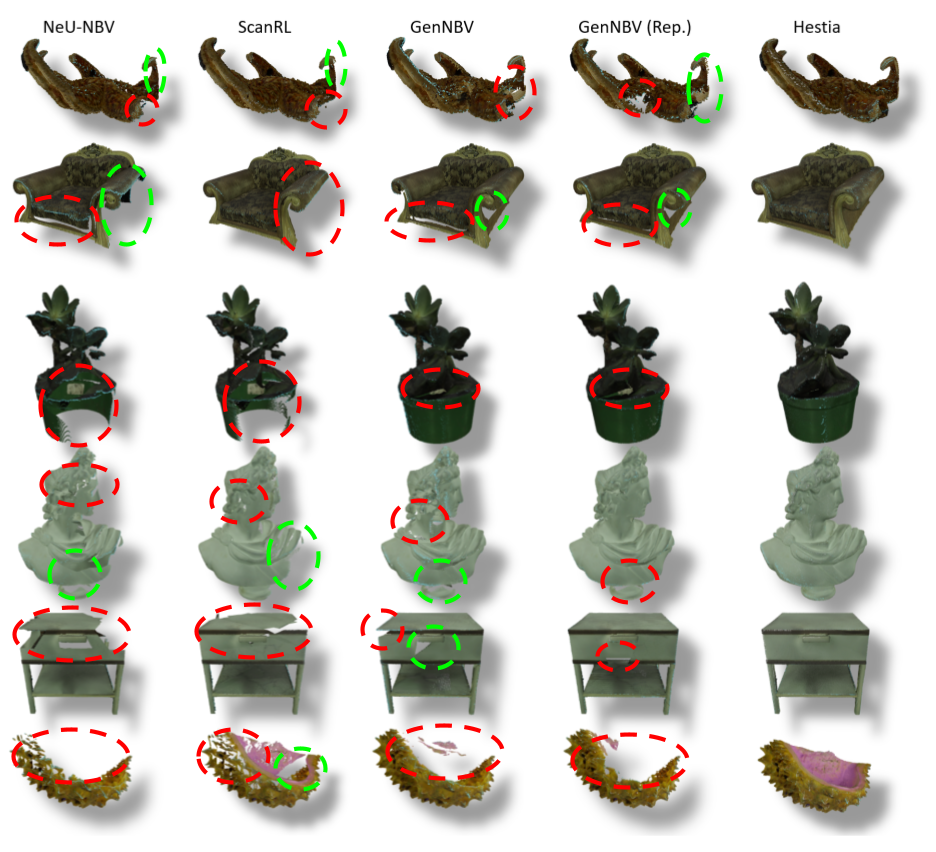}
  \caption{\textbf{Qualitative comparison on OmniObject3D~\cite{wu2023omniobject3d}.} Hestia's viewpoints reconstruct the point clouds of real-world scanned objects accurately, while other baselines exhibit less robustness across various object shapes, often missing parts in the reconstructed point clouds.}
  \label{afig:omni_all}
\end{figure*}

\begin{figure*}[tbh!]
  \centering
  \includegraphics[width=\linewidth]{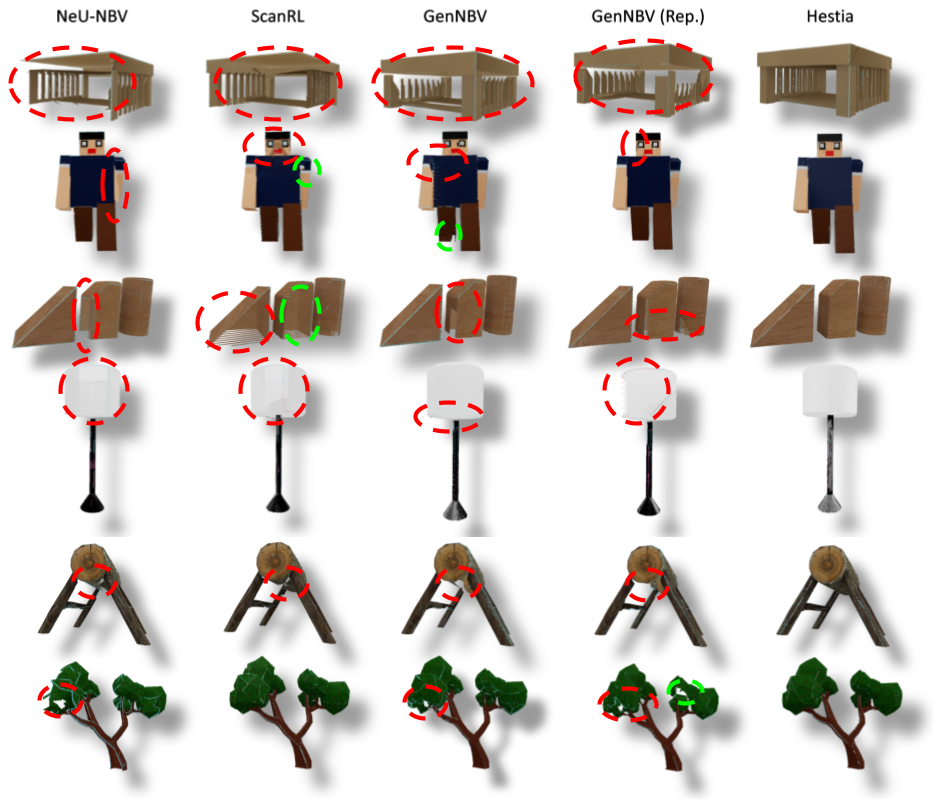}
  \caption{\textbf{Qualitative comparison on Objaverse~\cite{Deitke_2023_CVPR}.} Hestia’s viewpoints successfully reconstruct diverse and complex object shapes, while other baselines exhibit less robustness, often missing self-occluded regions or parts that require bottom-up viewpoints.}
  \label{afig:obj_all}
\end{figure*}

\begin{figure*}[tbh!]
  \centering
  \includegraphics[width=\linewidth]{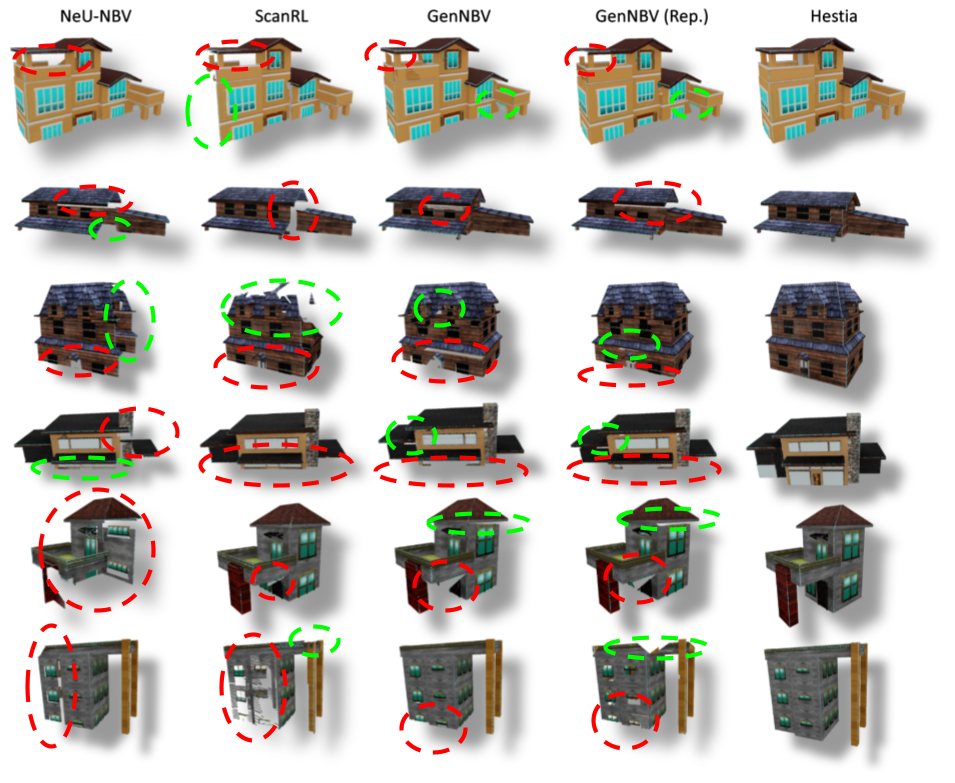}
  \caption{\textbf{Qualitative comparison on Houses3K~\cite{peralta2020next}.} Compared to the baselines, the point clouds reconstructed from the depth maps collected by Hestia capture finer details, such as roof soffits, pillars, and windows, particularly in self-occluded areas.}
  \label{afig:h3k_all}
\end{figure*}

\begin{figure*}[tbh!]
  \centering
  \includegraphics[width=\linewidth]{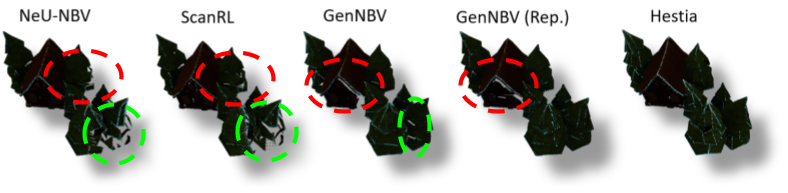}
  \caption{\textbf{Reconstruction on a complex scene.} Hestia captures the scene well.}
  \label{fig:vis_all2}
\end{figure*}

\begin{figure*}[tbh!]
  \centering
  \includegraphics[width=\linewidth]{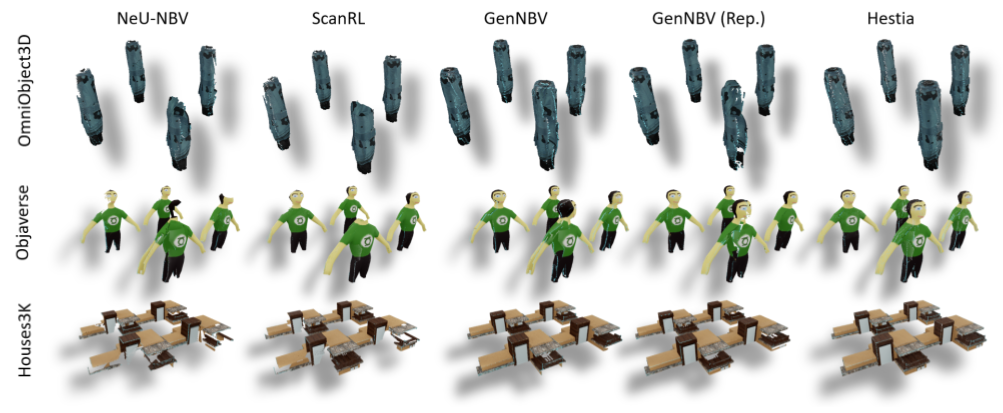}
  \caption{\textbf{Qualitative comparison of objects at the four corners.} Hestia performs robustly across different object position settings, while other baselines fail to maintain consistent performance across positions.}
  \label{afig:shift_all}
\end{figure*}

In this section, we present additional qualitative results for OmniObject3D~\cite{wu2023omniobject3d}, Objaverse~\cite{Deitke_2023_CVPR}, and Houses3K~\cite{peralta2020next} in~\cref{afig:omni_all,afig:obj_all,afig:h3k_all,fig:vis_all,afig:shift_all}.
\cref{afig:omni_all} shows that Hestia’s viewpoints successfully reconstruct the point clouds of real-world scanned objects, while other baselines often miss parts and perform less robustly across various object shapes. 
Specifically, other baselines miss the starfish’s arms, the sofa’s front or bottom, the plant’s pot, the statue’s head or stand, the table’s surface, and the durian’s flesh.
All these diverse missing parts are captured by Hestia.
\cref{afig:obj_all} shows that Hestia’s viewpoints robustly cover various object shapes, while other baselines often miss finer details or parts underneath. 
Specifically, other methods miss the wooden stand’s legs, the Lego man's face or arms, the lamp’s lampshade or neck, the underside of the wooden log, and the tree’s leaves. 
In contrast, Hestia successfully captures all these diverse and challenging parts.
\cref{afig:h3k_all} shows that Hestia’s viewpoints can effectively capture building-like structures, while other methods often miss features such as pillars, roof soffits, or windows.
\cref{fig:vis_all2} shows that Hestia can capture the complex scene well.
\cref{afig:shift_all} shows that Hestia is the only method that achieves consistent performance across different object position settings.
These visualization results validate that our proposed core components, including dataset choice, observation design, action space, reward calculation, and learning scheme, form a significant foundation for the tasks and thereby bring a non-marginal impact.

\section{Real-World Results}
\label{asec:realworld_res}

\begin{figure*}[tbh!]
  \centering
  \includegraphics[width=\linewidth]{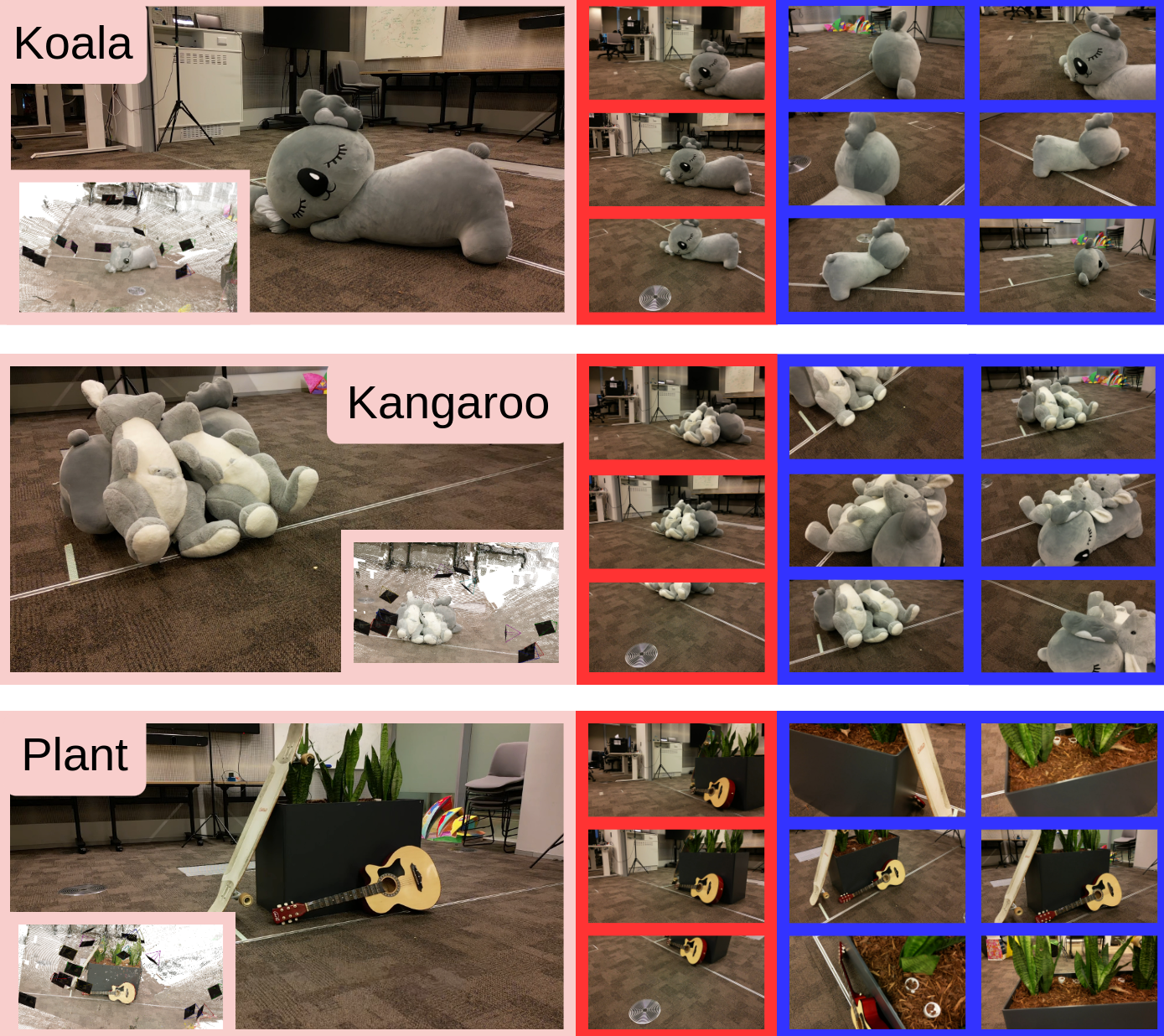}
  \caption{\textbf{Real-world demonstration of various object-centric scenes.} Hestia operates in real-world object-centric scenes starting from three initial viewpoints (red rectangles) and predicts the next-best viewpoints for capture (blue rectangles for the first six views), even when the depth camera is unavailable. Point cloud reconstruction results are shown on the left, with black rectangles representing the camera poses.}
  \label{afig:real_world2}
\end{figure*}

In this section, we present real-world images captured using Hestia operating within the drone system (see~\cref{asec:realworld_sys}).
\cref{fig:vis_real} demonstrates that Hestia performs well in real-world object-centric scenes, even when the depth camera is unavailable, for both shifted and non-shifted cases. 
Notably, without a depth camera to synchronize the virtual and real world, we manually set up three viewpoints, which are deliberately placed close to each other (e.g., the red boxes in~\cref{fig:vis_real}). 
In addition, it is reasonable that some next-best viewpoints appear similar because we use a multi-view depth predictor~\cite{duisterhof2024mast3r, wang2024dust3r} to convert RGB images into depth maps.
Therefore, it is common for certain viewpoints to overlap in order to obtain depth and update the input state.
\cref{afig:real_world2} shows that Hestia robustly handles various real-world object shapes. 
These results demonstrate that Hestia surpasses the prior works~\cite{guedon2023macarons, peralta2020next, chen2024gennbv, jin2023neu, lin2022active, sunderhauf2023density, pan2022activenerf, ran2023neurar, zhan2022activermap}, which have not been validated in real-world environments. 
In addition, Hestia is suitable for use as a viewpoint predictor for real-world applications.

\section{Reward Design}
\label{asec:rew}

In this section, we review the design of our reward function.
The reward function in Hestia is formulated as
\begin{equation}
r_t = R(s_t, a_t) = r_{\text{coverage}}(s_t, a_t) + r_{\text{constraint}}(s_t, a_t).
\end{equation}
To promote the exploration of previously unseen surfaces, we define the positive reward as
\begin{align}
r_{\text{coverage}}(s_t, a_t) 
&= \frac{\sum_{i=1}^N \sum_{j=1}^6 \left( F_t^{i,j} - F_{t-1}^{i,j} \right) \cdot M_{\text{col}}}{N \cdot 6} \notag \\
&\quad \cdot 0.3
\end{align}
where $F_t^{i,j}$ and $F_{t-1}^{i,j}$ denote the visibility status of the $j$-th face of the $i$-th voxel at time $t$ and $t-1$, respectively. 
The variable $M_{\text{col}} \in \{0, 1\}$ acts as a collision indicator, set to $0$ when a collision occurs, thereby nullifying any potential reward for unsafe actions.
This reward is computed based on the increment in newly visible voxel faces at the current step. 
By focusing on the increment rather than the accumulated visibility, the agent is better able to associate rewards with the immediate effects of its actions.
To discourage unsafe or invalid decisions, we define a penalty as
\begin{equation}
r_{\text{constraint}}(s_t, a_t) =
\begin{cases} 
-0.01, & \text{if } r_{\text{coverage}}(s_t, a_t) = 0, \\
       & \text{or } a_t[2] > H_t, \\
       & \text{or } a_t \in \text{non-free voxels}, \\
0,     & \text{otherwise.}
\end{cases}
\end{equation}
In particular, a negative reward is assigned if the agent fails to reveal any new faces, attempts to move above the maximum allowable flight height $H_t$, or selects a viewpoint located within non-free voxels.
To ensure a balance between positive and negative rewards, the positive reward is scaled by a factor of $0.3$.  
This weighting is based on the observation that the maximum face ratio is 1 and each episode ends after 50 steps.  
As a result, the total possible positive reward is approximately $0.3 \times 1 = 0.3$, which roughly aligns with the maximum overall value of the negative penalty $0.01 \times 50 = 0.5$.  
If the episode ends earlier, such as after 30 steps, this reward structure maintains a perfect balance.

\section{Novelty Justification}
\label{asec:novel}

\begin{table*}[tbh!]
    \centering
\begin{tabular}{
    p{2.5cm}| 
    >{\centering\arraybackslash}p{1.0cm}
    >{\centering\arraybackslash}p{1.0cm}
    >{\centering\arraybackslash}p{1.0cm}
    >{\centering\arraybackslash}p{0.8cm}
    >{\centering\arraybackslash}p{1.0cm}
    >{\centering\arraybackslash}p{0.7cm}
    >{\centering\arraybackslash}p{0.9cm}
    >{\centering\arraybackslash}p{0.8cm}
}
        \multirow{3}{*}{Method} & No Cand. View & No Online Learn. & Voxel as Cube & Hier. Str. & Greedy & Div. Data. & Robust for Shift. & Real-World Demo. \\
        \specialrule{.1em}{.1em}{.1em} 
        Active3D~\cite{lin2022active} & \xmark & \xmark & \NA & \xmark & \NA & \xmark & \NA & \xmark \\
        NeRF-En~\cite{sunderhauf2023density} & \xmark & \xmark & \NA & \xmark & \NA & \xmark & \NA & \xmark \\
        ActiveNeRF~\cite{pan2022activenerf} & \xmark & \xmark & \NA & \xmark & \NA & \xmark & \NA & \xmark \\
        NeurAR~\cite{ran2023neurar} & \xmark & \xmark & \NA & \xmark & \NA & \xmark & \NA & \xmark \\
        EfficientView~\cite{zeng2022efficient} & \xmark & \xmark & \NA & \xmark & \NA & \xmark & \NA & \checkmark \\
        UnGuide~\cite{lee2022uncertainty} & \xmark & \xmark & \NA & \xmark & \NA & \xmark & \NA & \checkmark \\
        SEER~\cite{tao2022seer} & \xmark & \xmark & \NA & \checkmark & \NA & \xmark & \NA & \checkmark \\
        ActiveRMAP~\cite{zhan2022activermap} & \xmark & \xmark & \NA & \xmark & \NA & \xmark & \NA & \xmark \\
        MACARONS~\cite{guedon2023macarons} & \xmark & \xmark & \NA & \xmark & \NA & \NA & \NA & \xmark \\
        \specialrule{.1em}{.1em}{.1em} 
        NeU-NBV~\cite{jin2023neu} & \xmark & \checkmark & \NA & \xmark & \NA & \xmark & \xmark & \xmark \\
        ScanRL~\cite{peralta2020next} & \checkmark & \checkmark & \NA & \xmark & \xmark & \xmark & \xmark & \xmark \\
        GenNBV~\cite{chen2024gennbv} & \checkmark & \checkmark & \xmark & \xmark & \xmark & \xmark & \xmark & \xmark \\
        \specialrule{.1em}{.1em}{.1em} 
        Hestia & \checkmark & \checkmark & \checkmark & \checkmark & \checkmark & \checkmark & \checkmark & \checkmark \\
        \specialrule{.1em}{.1em}{.1em}
    \end{tabular}
    \caption{\textbf{Comparison of learning-based next-best-view methods.} Compared to online learning methods, Hestia achieves real-time inference speed. Compared to generalizable methods that predict five-degree-of-freedom viewpoints or assume a drone as the agent, Hestia exhibits robustness across different object position settings and demonstrates feasibility in real-world object-centric scenes.}
    \label{atab:nbv_comparison}
\end{table*}
This section elaborates on the novelty of Hestia.
Hestia is a generalizable next-best-view planner that can predict five-degree-of-freedom viewpoints and model a drone as an agent.
Therefore, we mainly focus on comparing methods that also predict five-degree-of-freedom viewpoints or assume a drone as an agent.
Unlike prior approaches, Hestia systematically addresses the next-best-view task by introducing core components such as dataset selection, observation design, action space formulation, reward computation, and learning schemes. 
Together, these elements form a comprehensive and unified foundation for the planner (see~\cref{sec:abl}).
As shown in~\cref{atab:nbv_comparison}, compared to online-learning methods~\cite{lin2022active, sunderhauf2023density, pan2022activenerf, ran2023neurar, lee2022uncertainty, zhan2022activermap, guedon2023macarons}, Hestia avoids the need to sample candidate views or perform online optimization. 
This results in greater flexibility in viewpoint prediction and supports real-time inference.
Additionally, in comparison to generalizable methods~\cite{jin2023neu, peralta2020next, chen2024gennbv}, Hestia is trained on a significantly larger and more diverse dataset, enabling it to generalize robustly to a wide variety of object shapes during testing. 
Hestia is also the only method that consistently performs well under different object configurations, as demonstrated in~\cref{atab:base_bench_omni,atab:base_bench_obj,atab:base_bench_h3k}.
These advantages stem from the key innovations in our design, including treating voxels as cubes rather than points, employing a hierarchical structure to manage the complexity of the action space, and using a greedy learning scheme to mitigate spurious correlations.
Notably, the purpose of Hestia’s hierarchical structure is to address the high-dimensional continuous action search space problem in reinforcement learning-based generalizable next-best-view planning, which is fundamentally different from traditional methods that use hierarchical structures to move along frontiers. 
One of the most recent works~\cite{chen2024gennbv} still lacks the designs we propose.

\section{Spurious Correlation}
\label{asec:spur}

\begin{figure*}[tbh!]
   \centering
\includegraphics[width=0.95\linewidth]{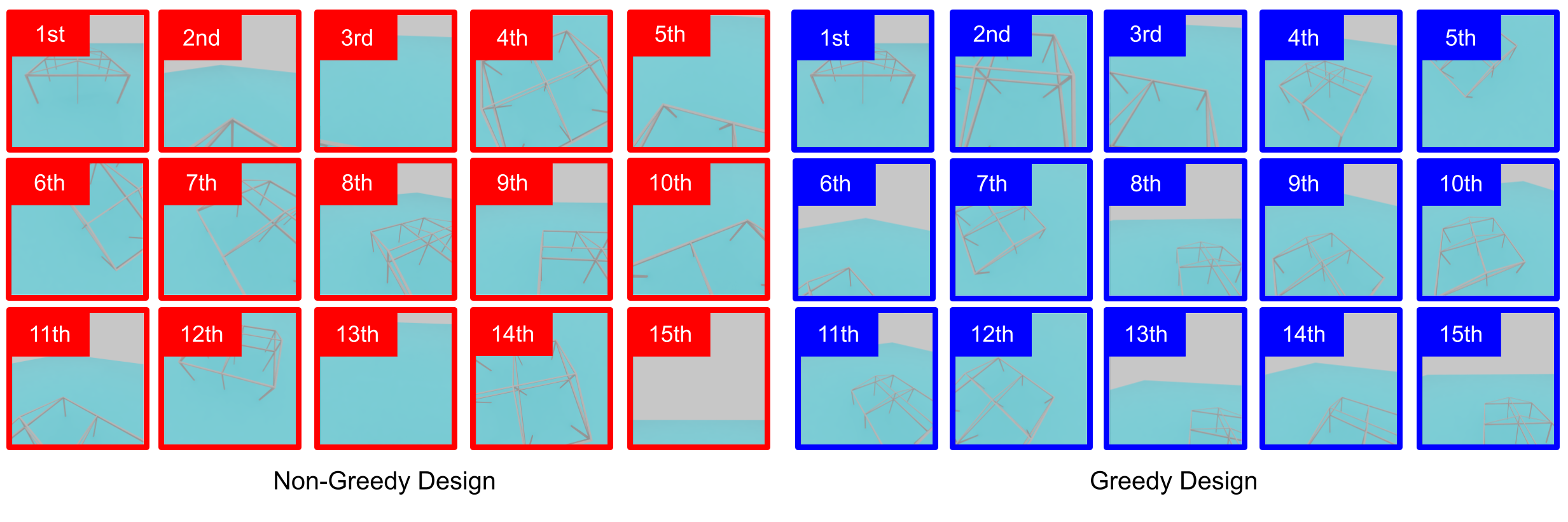}
   \caption{\textbf{Spurious correlation.} In this study, spurious correlation refers to the assignment of future positive rewards to current non-beneficial actions, resulting in suboptimal viewpoint predictions. For instance, the third, thirteenth, and fifteenth viewpoints are empty in the non-greedy design, while enabling the close-greedy design alleviates this issue.}
   \label{fig:spur_cmp}
\end{figure*}

In this section, we present the spurious correlation caused by future positive rewards in the task.
Spurious correlation has been widely observed across various tasks~\cite{interpretation1971spurious, kim2024discovering}.  
In our task, we find that using a large discount factor and future goal rewards can lead to false associations between current actions and their rewards.  
This creates an illusion for the reinforcement learning agent that the current action is beneficial, even when there is no information gain (e.g., empty views as shown in~\cref{fig:spur_cmp}) resulting from the current action.
Enabling the close-greedy design mitigates this issue as shown in~\cref{fig:spur_cmp}.

\section{Datasets}
\label{asec:datasets}

This section briefly introduces the three main datasets used in the Hestia benchmark: Houses3K~\cite{peralta2020next}, Objaverse~\cite{deitke2024objaverse, Deitke_2023_CVPR}, and OmniObject3D~\cite{wu2023omniobject3d}.

\begin{figure*}[tbh!]
  \centering
  \includegraphics[width=0.8\linewidth]{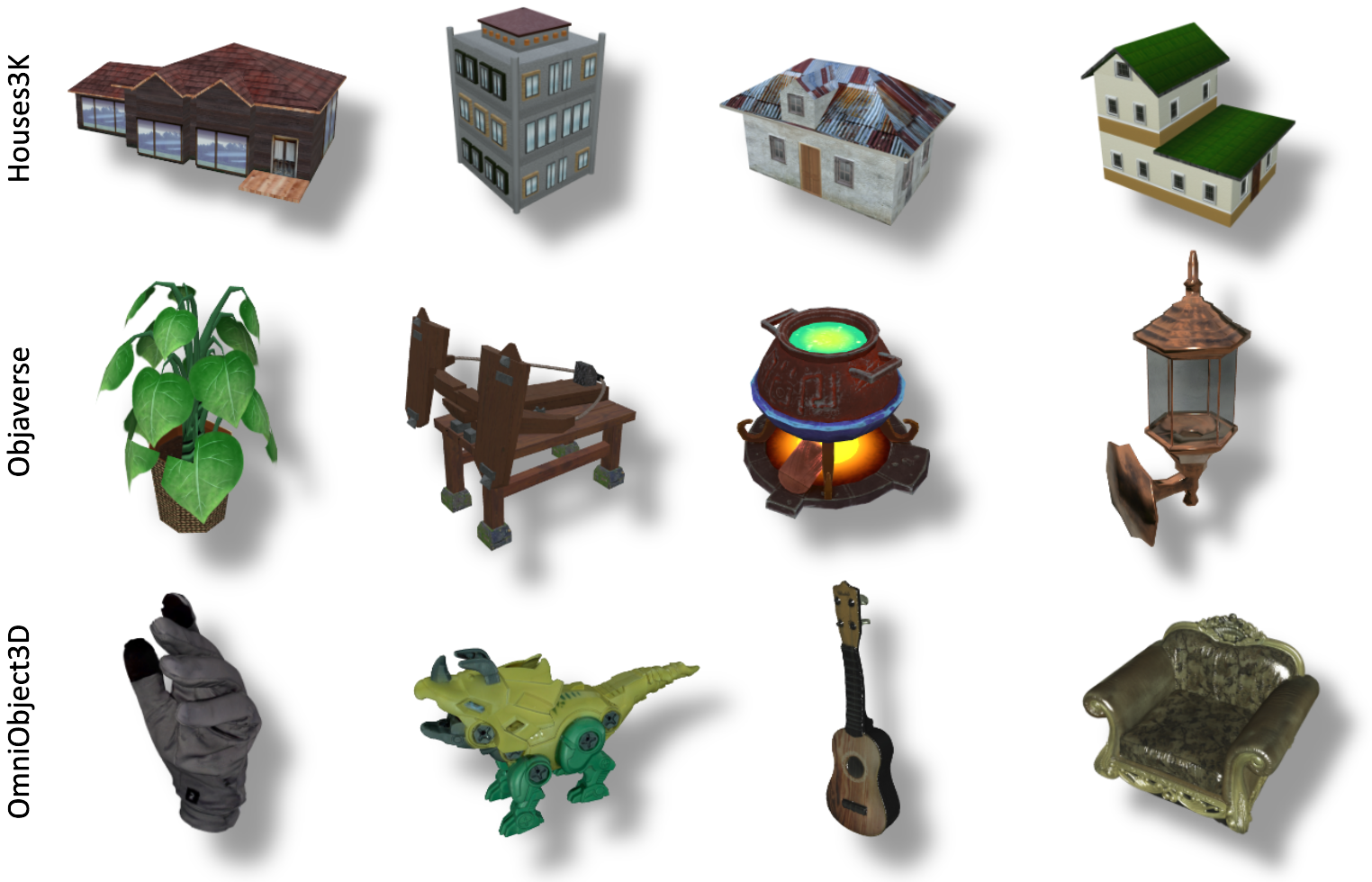}
  \caption{\textbf{Sample data from Houses3K~\cite{peralta2020next}, Objaverse~\cite{Deitke_2023_CVPR}, and OmniObject3D~\cite{wu2023omniobject3d}.}
  Houses3K features building shapes with challenging self-occlusions, such as roof soffits. Objaverse includes a diverse range of object shapes. OmniObject3D contains high-quality real-world 3D scans.
  }
  \label{afig:datasets_all}
\end{figure*}

\subsection{Introduction to Datasets}

\textbf{Houses3K Dataset.} Houses3K~\cite{peralta2020next} is designed for next-best-view policy learning.  
The dataset contains 600 distinct buildings, each rendered with five texture variants, yielding a total of 3,000 FBX models.  
Many buildings feature challenging self-occlusions, such as roof soffits, that can be fully observed only from bottom-up viewpoints (see~\cref{afig:datasets_all}). 
However, because the dataset includes only a single cube-like object category (e.g., buildings), its diversity is limited, which may hinder the ability of next-best-view policies trained on Houses3K to generalize to other structures or everyday objects.
\\
\textbf{Objaverse Dataset.} 
Objaverse~\cite{Deitke_2023_CVPR} is one of the largest open 3D datasets, containing more than 800,000 shapes across at least 18 high-level categories, including furniture, vehicles, animals, and plants (see~\cref{afig:datasets_all}).   
Each category is further divided into several subcategories.  
The dataset’s scale and diversity make it particularly well-suited for foundation model research, especially for 3D generative models.  
To the best of our knowledge, we are the first to introduce Objaverse for next-best-view policy learning.  
Its large-scale and diverse object coverage enables training next-best-view policies that perform robustly across a wide range of categories and shapes.
\\
\textbf{OmniObject3D Dataset.}
OmniObject3D~\cite{wu2023omniobject3d} is a high-quality 3D object dataset collected through real-world scanning, consisting of approximately 6,000 objects across more than 190 categories (see~\cref{afig:datasets_all}).  
Unlike synthetic datasets, OmniObject3D captures real-world geometry and texture details using high-resolution 2D and 3D sensors.  
It provides accurate geometry and realistic material properties, making it commonly used for evaluating real-world transferability in vision tasks such as novel-view synthesis.  
In this paper, we introduce OmniObject3D for benchmark purposes.

\subsection{Dataset Preparation}
\begin{figure*}[tbh!]
  \centering
  \includegraphics[width=0.8\linewidth]{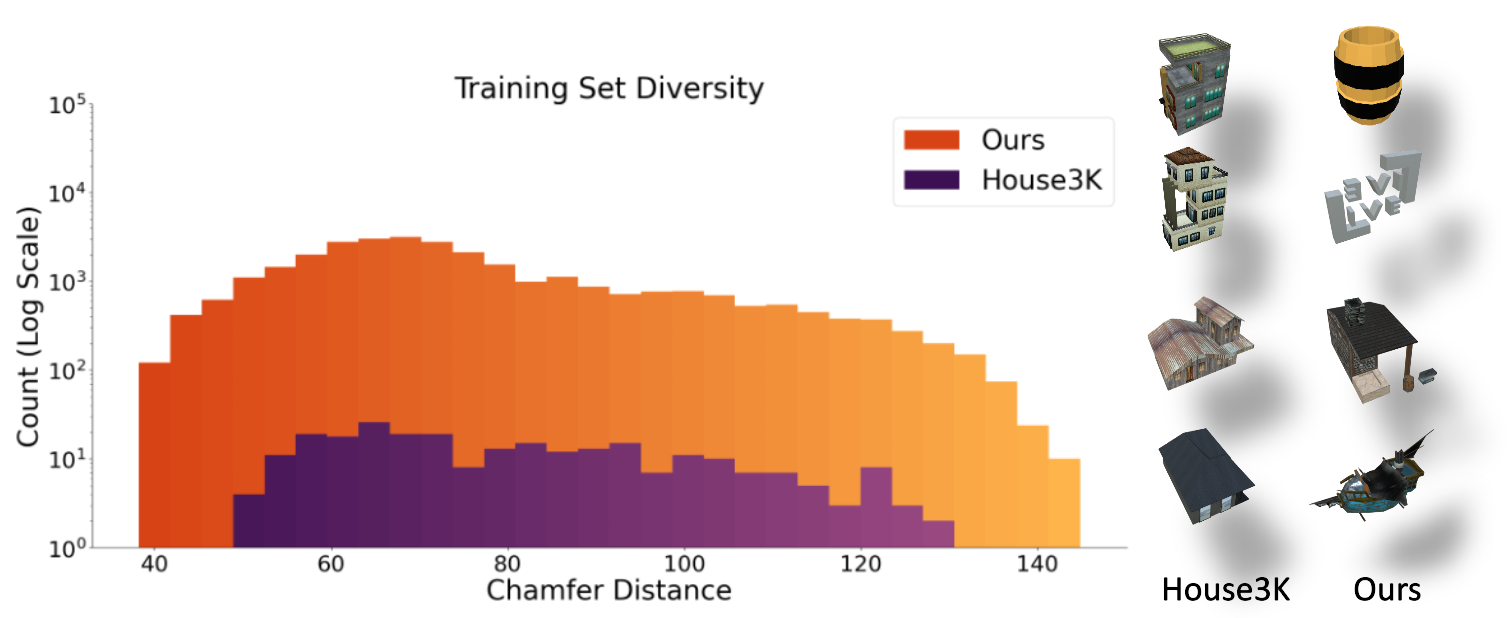}
  \caption{\textbf{More diverse and large-scale training set.} The chamfer distance measures the discrepancy between the point cloud of the training data and the sphere point cloud.  
  The logarithmic scale of the count represents the number of shapes within each distance range.  
  The right portion displays sample shapes with the same chamfer distance, shown side by side for each dataset. 
  The wider chamfer distance range, higher number of shapes per chunk, and varied shape categories demonstrate that our training data, processed from Objaverse~\cite{deitke2024objaverse, Deitke_2023_CVPR}, are more comprehensive and large-scale compared to Houses3K.
  }
  \label{fig:datasets}
\end{figure*}
\begin{figure*}[tbh!]
  \centering
  \includegraphics[width=0.8\linewidth]{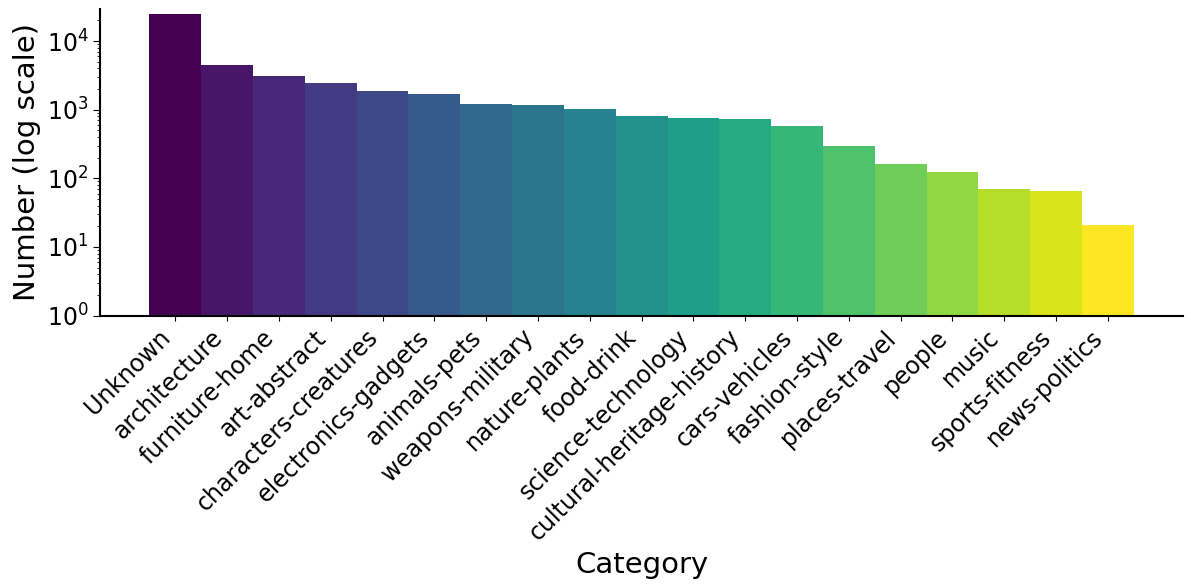}
  \caption{\textbf{Dataset distribution.} Our training dataset, processed from Objaverse~\cite{deitke2024objaverse, Deitke_2023_CVPR}, includes a wide range of categories and is not limited to cubic-like shapes (e.g., buildings).
  }
  \label{fig:data_dis}
\end{figure*}

We propose using Objaverse~\cite{deitke2024objaverse, Deitke_2023_CVPR} as the training dataset to ensure a diverse range of shapes during training (see~\cref{fig:datasets,fig:data_dis}).
To achieve this, we filter out large meshes and download the remaining mesh files from Objaverse~\cite{deitke2024objaverse, Deitke_2023_CVPR}, resulting in a dataset comprising 120,000 shapes.  
For each shape, we generate the occupancy grid and point cloud using Open3D~\cite{zhou2018open3d}.  
To remove invisible voxels and points, we perform a breadth-first search (BFS) starting from external free voxels, retaining only reachable occupancy voxels and points as the ground truth.  
The visible faces of each voxel are identified by examining the occupancy states of neighboring voxels.  
Unsupervised learning~\cite{yang2024toward,fei2025discovering,DBLP:conf/aaai/YangLWCXZDLJ26,macqueen1967some} is efficient for data selection, where we apply PCA~\cite{abdi2010principal} to reduce the point clouds to three components and use k-means clustering~\cite{macqueen1967some} to group the 120,000 shapes into 30,100 clusters.
The cluster centers of 30,000 clusters are designated as training data, while the remaining cluster centers are used for testing.  
The same procedure is used to create 256 training samples and 100 test samples from the Houses3K dataset~\cite{peralta2020next} for our benchmark.  
Our processed training set is two orders of magnitude larger than those used in previous studies~\cite{jin2023neu, chen2024gennbv, peralta2020next} and includes at least 18 more categories than prior datasets~\cite{chen2024gennbv, peralta2020next}.
As for OmniObject3D~\cite{wu2023omniobject3d}, due to limited storage space, we randomly select one shape per category for evaluation, resulting in approximately 200 test samples.

\begin{figure*}[tbh!]
  \includegraphics[width=\linewidth]{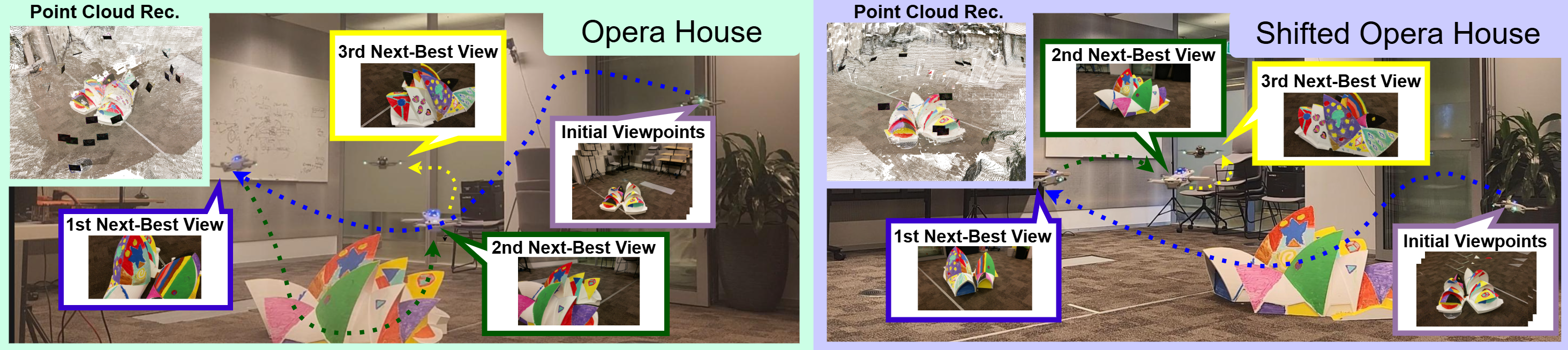}
  \captionof{figure}{\textbf{Real-world drone system.} Hestia is a generalizable next-best-view planner that is feasible for real-world deployment.
  \textit{Please refer to our demonstration video for further details.}}
  \label{fig:teaser}
\end{figure*}
\label{sec:realworld_sys}
\begin{figure*}[tbh!]
  \centering
  \includegraphics[width=\linewidth]{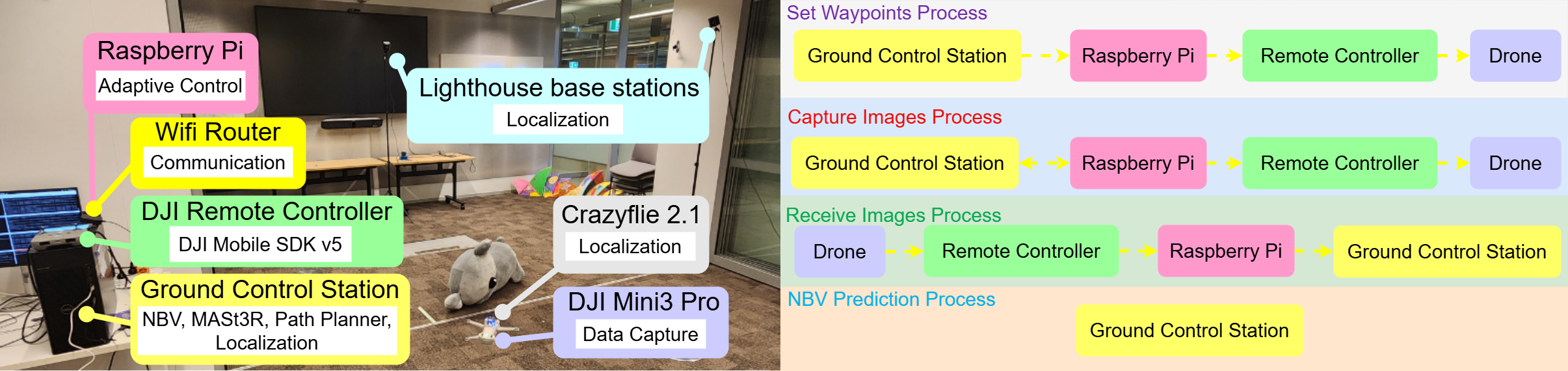}
  \caption{\textbf{Overview of the real-world drone system.} The system uses a drone with an RGB camera for data capture, Lighthouse base stations and Crazyflie for localization, and a Wifi router for wireless communication.}
  \label{fig:real_world}
\end{figure*}

\section{Real-World Drone System}
\label{asec:realworld_sys}

This section includes the setup and pseudo code of the real-world drone system we used.
By integrating Hestia into the real-world drone system (see~\cref{fig:teaser}), Hestia demonstrates its feasibility under practical scenarios.  

\begin{figure*}[tbh!]
\centering
\begin{minipage}{\linewidth}
\begin{algorithm}[H]
\caption{\textcolor[HTML]{D2EBF0}{Main}}
\label{alg:main}
\begin{algorithmic}[1]
\Require A drone system 
$\mathcal{D} = \{\mathcal{D}_\text{gs}, \mathcal{D}_\text{pi}, \mathcal{D}_\text{ad}, \mathcal{D}_\text{ct}, \mathcal{D}_\text{drone}\}$, \textbf{where:}
\begin{flushleft}
\hspace{1em} $\mathcal{D}_\text{gs}$: Ground station \\
\hspace{1em} $\mathcal{D}_\text{pi}$: Raspberry Pi \\
\hspace{1em} $\mathcal{D}_\text{ad}$: Android phone \\
\hspace{1em} $\mathcal{D}_\text{ct}$: Remote controller \\
\hspace{1em} $\mathcal{D}_\text{drone}$: Drone
\end{flushleft}

\State $\mathcal{X} \gets [x_1, x_2, x_3]$ \Comment{Initial viewpoints}
\For{$x \in \mathcal{X}$}
    \State $\mathcal{W} \gets [x]$ \Comment{Update waypoints}
    \State \textcolor[HTML]{682C99}{\texttt{Set\_Waypoints($\mathcal{W}$)}} \Comment{\textcolor[HTML]{682C99}{Move the drone (Alg. 2)}}
    \State \textcolor[HTML]{D00402}{$i \gets \texttt{Capture\_Image()}$} \Comment{\textcolor[HTML]{D00402}{Capture an image (Alg. 3)}}
    \State \textcolor[HTML]{069F48}{\texttt{Receive\_Image(i)}} \Comment{\textcolor[HTML]{069F48}{Transmit image (Alg. 4)}}
\EndFor

\For{$k \in \{1, 2, \dots, K\}$}
    \State \textcolor[HTML]{06AFF5}{$\mathcal{W} \gets \texttt{NBV\_Prediction()}$} \Comment{\textcolor[HTML]{06AFF5}{Predict NBV (Alg. 5)}}
    \State \textcolor[HTML]{682C99}{\texttt{Set\_Waypoints($\mathcal{W}$)}} \Comment{\textcolor[HTML]{682C99}{Move to NBV (Alg. 2)}}
    \State \textcolor[HTML]{D00402}{$i \gets \texttt{Capture\_Image()}$} \Comment{\textcolor[HTML]{D00402}{Capture an image (Alg. 3)}}
    \State \textcolor[HTML]{069F48}{\texttt{Receive\_Image(i)}} \Comment{\textcolor[HTML]{069F48}{Transmit image (Alg. 4)}}
\EndFor

\end{algorithmic}
\end{algorithm}
\end{minipage}
\end{figure*}

\subsection{Real-World System Overview}

To demonstrate Hestia's feasibility in a real-world environment, we use a real-world system (see~\cref{fig:real_world}), where a drone equipped with an RGB camera moves to the next-best viewpoint predicted by Hestia to capture images of an object. 
The system uses four HTC Lighthouse base stations and a Crazyflie 2.1 for localization and transmits images to the ground control station via wireless communication.
MASt3R~\cite{duisterhof2024mast3r} is integrated to convert RGB images into pointmaps (e.g., depth images), and three initial viewpoints are set for real-world and virtual-world synchronization~\cite{umeyama1991least}. 
\cref{fig:real_world} and Alg.~\ref{alg:main} illustrate four key processes of the system. 
Specifically, the drone captures images at three initial viewpoints, where the \textit{set waypoints process} navigates the drone using a heuristic trajectory planner based on prior knowledge of the environment. 
Then, the \textit{capture image process} commands image capture, and the \textit{receive image process} transmits the image to the ground station. 
After capturing the initial viewpoints, the \textit{nbv prediction process} predicts the next-best viewpoint based on the collected data.
Four processes repeat until sufficient data is collected. 
For more details, please refer to~\cref{asec:process}.

\subsection{Real-World System Setup}

The environment size of our object-centric scenes (e.g., an opera house) is approximately 2.6\,m\,$\times$\,2.6\,m\,$\times$\,2\,m.  
To prevent the drone from exceeding the HTC Lighthouse base station range, the maximum height $H_t$ is restricted to 1.5\,m.
Additionally, in the nearest collision-free voxel module, voxels below 0.4m are marked as occupied to avoid potential counterforces between the floor and the drone's quadrotor.
In this system, we deploy the DJI Mini 3 Pro as the primary aircraft model. 

We integrate the DJI Mobile SDK v5 to enable remote control and command transmission from the base station to the UAV.  
This software development kit provides developers with comprehensive control capabilities over the UAV.  
The SDK is embedded in an Android application package, where we develop a custom application capable of broadcasting aircraft data via the User Datagram Protocol (UDP) wireless network protocol.  
The broadcast data is captured using a Raspberry Pi 4, which runs a ROS 2 node designed to receive UDP packets and convert them into ROS 2-compatible messages.  
On the same Raspberry Pi 4, we implement an adaptive trajectory planning technique.  
This method evaluates a pre-generated library of feasible offline trajectories, allowing the UAV to navigate autonomously by selecting the most appropriate path based on real-time conditions.  
Integrating these components ensures a reliable flow of data and commands, enabling efficient autonomous navigation for the UAV.
We utilize the Crazyflie v2.1, a nano UAV equipped with a Lighthouse Positioning Deck, to achieve precise localization within the experimental environment.  
To integrate its capabilities with the primary aircraft, we remove the propellers and motors of the Crazyflie and securely mount it on top of the DJI Mini 3 Pro.  
This setup enables the Crazyflie to serve as a localization beacon, providing accurate positional data for the main UAV within the tracking range of the Lighthouse base stations.  
The positioning deck on the Crazyflie captures localization data using infrared signals from the Lighthouse system.  
This data is transmitted wirelessly via the Crazyradio 2.0 module, which connects to a base station.  
The base station, running the Crazyswarm 2.0 package on the ROS 2 framework, processes and publishes the localization data in real time.  
This setup facilitates autonomous navigation and precise positioning of the main UAV by continuously updating its coordinates within the experimental space.

The system is constructed entirely from publicly available, low-cost hardware.  
The total cost of the additional components, including the Crazyflie v2.1 (approximately \$200), the Lighthouse Positioning Deck (around \$100), and the Crazyradio 2.0 module (about \$50), is approximately \$350.  
When combined with the DJI Mini 3 Pro, which costs around \$800, the total system cost remains significantly lower than that of conventional localization solutions.  
This cost-effective design, combined with open-source software such as ROS 2 and the Crazyswarm package, provides a reliable and accessible prototype for UAV-based data collection.

\subsection{Real-World System Processes}
\label{asec:process}

The flowchart of the system in the right part of~\cref{fig:real_world}, highlights four main sub-processes: \textit{Set Waypoints Process}, \textit{Capture Image Process}, \textit{Receive Image Process}, and \textit{NBV Prediction Process}. 
Each sub-process is represented by distinct colors in the diagram and described as follows:
\begin{itemize}
    \item \textbf{Setting Waypoints (Alg.~\ref{alg:set_w})}: Waypoints for the drone are pre-configured and stored on the ground station. These waypoints are transmitted to the drone through a communication channel comprising a Raspberry Pi, an Android mobile phone connected to the remote controller, and a remote controller. The drone navigates to each waypoint to align with the predicted NBV.

\begin{algorithm}[H]
\caption{\textcolor[HTML]{682C99}{Set Waypoints}}
\label{alg:set_w}
\begin{algorithmic}[1]
\Require Waypoints $\mathcal{W}$, and a drone system $\mathcal{D}$

\State $\mathcal{D}_\text{gs} 
\xrightarrow{\mathcal{W}} \mathcal{D}_\text{pi}
\xrightarrow{\mathcal{W}} \mathcal{D}_\text{ad}
\xrightarrow{\mathcal{W}} \mathcal{D}_\text{ct}
\xrightarrow{\mathcal{W}} \mathcal{D}_\text{drone}$
\Comment{Send waypoints to the drone}

\For{$w \in \mathcal{W}$}
    \State $\mathcal{D}_\text{drone}.\texttt{move\_to}(w)$ \Comment{Move to waypoint}
\EndFor

\end{algorithmic}
\end{algorithm}

    \item \textbf{Capturing Images (Alg.~\ref{alg:cap_img})}: Upon reaching a waypoint, the Raspberry Pi retrieves the drone's real-time position from the ground station. Once the waypoint is confirmed, the ground station sends a "capture image" command. The drone then captures the image using adjusted camera parameters.

\begin{algorithm}[H]
\caption{\textcolor[HTML]{D00402}{Capture Image}}
\label{alg:cap_img}
\begin{algorithmic}[1]
\Require A drone system $\mathcal{D}$

\While{$\mathcal{D}_\text{drone}.\text{loc} \not\approx \mathcal{D}_\text{pi}.\text{x}$}
    \Comment{Wait until location matches NBV}
    \State Continue
\EndWhile

\State $\mathcal{D}_\text{pi} \xrightarrow{\text{NBV\_reached}} \mathcal{D}_\text{gs}$ \Comment{Notify ground station}
\State $i \gets \mathcal{D}_\text{drone}.\texttt{capture()}$
\Comment{Capture image}
\State \Return $i$ \Comment{Return image}

\end{algorithmic}
\end{algorithm}
    
    \item \textbf{Receiving Images (Alg.~\ref{alg:rec_img})}: After capturing an image, the drone transmits the image along with its real-time pose back to the ground station. These images are used to iteratively update the reconstruction model.

\begin{algorithm}[H]
\caption{\textcolor[HTML]{069F48}{Receive Image}}
\label{alg:rec_img}
\begin{algorithmic}[1]
\Require Image $i$ and a drone system $\mathcal{D}$

\State $\mathcal{D}_\text{drone} 
\xrightarrow{i} \mathcal{D}_\text{ct} 
\xrightarrow{i} \mathcal{D}_\text{ad}
\xrightarrow{i} \mathcal{D}_\text{pi}
\xrightarrow{i} \mathcal{D}_\text{gs}$
\Comment{Transmit image to ground station}

\State $\mathcal{D}_\text{gs}.\texttt{save}(i)$ \Comment{Save image}
\State $\mathcal{D}_\text{gs}.\texttt{save}(x')$ \Comment{Save real-time position}

\end{algorithmic}
\end{algorithm}

    \item \textbf{Predicting the NBV (Alg.~\ref{alg:nbv_pred})}: The ground station processes the captured image and the drone’s pose to predict the next-best-view using the NBV module. The newly determined viewpoint is then sent to the drone to continue the data collection process.

\begin{algorithm}[H]
\caption{\textcolor[HTML]{06AFF5}{NBV Prediction}}
\label{alg:nbv_pred}
\begin{algorithmic}[1]
\Require A drone system $\mathcal{D}$

\State $\mathcal{I} \gets [i_1, \dots, i_n]$ \Comment{Load images}
\State $\mathcal{X}^{'} \gets [x^{'}_{1}, \dots, x^{'}_{n}]$ \Comment{Load positions}
\State $\mathcal{G} \gets \mathcal{D}_\text{gs}.\texttt{MASt3R}(\mathcal{X}', \mathcal{I})$ \Comment{Compute grid}
\State $x \gets \mathcal{D}_\text{gs}.\texttt{pred\_NBV}(\mathcal{G}, i_n, x^{'}_{n}, h)$ \Comment{Predict NBV}
\State $\mathcal{W} \gets \mathcal{D}_\text{gs}.\texttt{generate\_waypoints}(x)$ \Comment{Generate waypoints}
\State \Return $\mathcal{W}$ \Comment{Return waypoints}

\end{algorithmic}
\end{algorithm}
\end{itemize}

Alg.~\ref{alg:main} provides an overview of the entire process. 
Initially, the drone visits three pre-defined viewpoints to synchronize the real-world and virtual-world data (lines 1-7). 
Following these initial captures, the NBV module predicts subsequent viewpoints based on the collected data (lines 8-13), guiding the drone iteratively until sufficient data is acquired for reconstruction.
Additionally, the MASt3R module is integrated into the ground station to convert RGB images into pointmaps (e.g., depth images). 
By combining these components, the system enables efficient and intelligent data collection, demonstrating the potential of drones as autonomous agents for scalable and versatile real-world scenarios.

\section{Training and Testing Details}
\label{asec:train_det}
\begin{figure*}[tbh!]
  \centering
  \includegraphics[width=\linewidth]{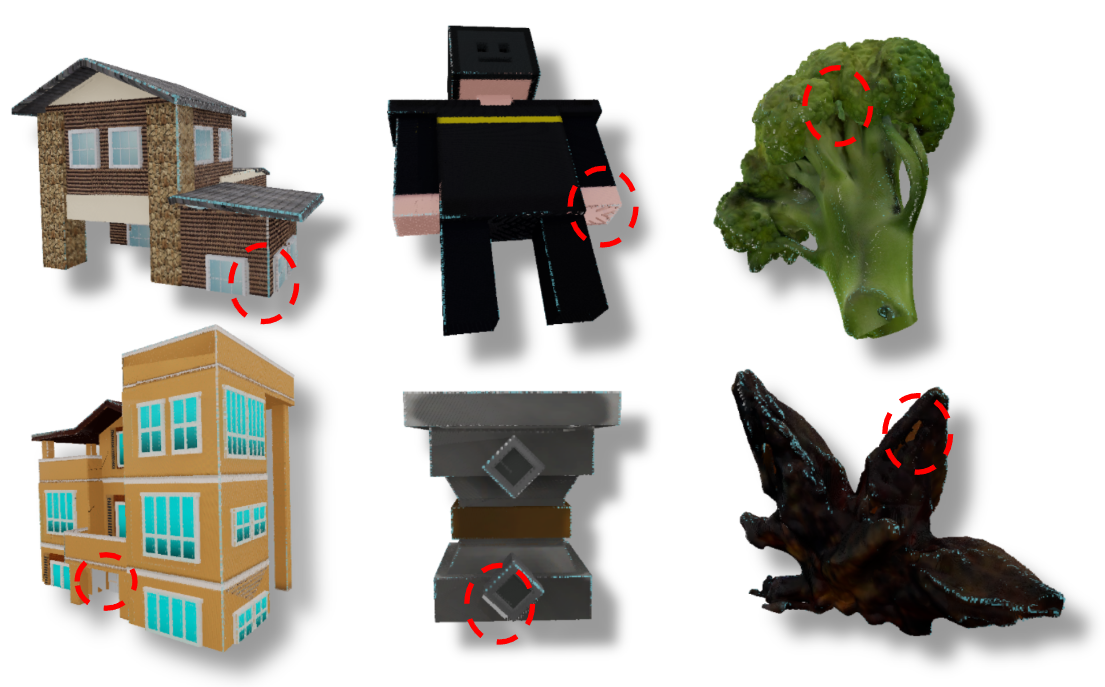}
  \caption{\textbf{Failure cases of Hestia.} Hestia may occasionally fail to capture finer 3D structures, highly
self-occluded parts, nearly vertical bottom-up views, and small details on coarse object surfaces.}
  \label{fig:fail}
\end{figure*}
This section includes the details of the training and testing.
We employ PPO~\cite{schulman2017proximal} from stable-baselines3~\cite{stable-baselines3} as the reinforcement learning framework for Hestia.  
The grid resolution $g$ is set to 20, and $h$ and $w$ are set to 300.  
The initial learning rate is $3 \times 10^{-4}$ and is decayed by a factor of 2 every 500,000 iterations starting from 2,000,000 iterations until reaching 4,000,000, for a total of 5,000,000 training iterations.  
During training, $H_t$ is randomly sampled from the initial viewpoint height, up to a maximum of 10 meters.  
For testing, $H_t$ starts at 10 meters, is reduced to 5 meters during the last 10 to 5 steps, and further decreases to 2 meters in the final 5 steps.  
A training episode ends and the scene resets either when the number of captured images reaches 50 or when the target face coverage ratio of 0.9 is achieved.  
The complete training process takes approximately 24 hours on an NVIDIA RTX A6000 GPU.

Our network architecture is lightweight, with only 4.9 million parameters (see~\cref{tab:ablation}, which is approximately half the size of a standard ResNet-18 model, which has around 11.7 million parameters.
The proposal network consists of three 3D convolutional layers, each followed by a Leaky ReLU activation.
This design progressively downsamples the 3D grid before further downstream operations.
It is then followed by a 3D self-attention layer, again paired with a Leaky ReLU activation, to expand the receptive field.
Finally, the network applies a reparameterization trick module, composed of linear layers, to generate the output distribution parameters.
This design allows the look-at point to be sampled from a distribution rather than predicted deterministically.
The grid encoder consists of three 3D convolutional layers, each followed by batch normalization and a Leaky ReLU activation.
After encoding, trilinear interpolation is applied to each encoded grid feature, followed by feature concatenation.
The image encoder is composed of three 2D convolutional layers, each followed by batch normalization and Leaky ReLU activation.
The encoded features are then flattened and passed through a linear layer with Leaky ReLU activation.
For the policy network, we adopt the default model provided in stable-baselines3.
For more details about the network architecture, please refer to the code provided in the supplementary materials.
The hierarchical design first predicts the look-at point, followed by the camera position.
This design prioritizes the look-at point, as the primary objective in this task is to determine where to look rather than where to fly.
It also resembles how a human pilot controls a drone during data capture, focusing first on the target of observation before planning the flight path.

\section{Limitations}
\label{asec:limit}

Although the quantitative and qualitative results (see ~\cref{sec:qua,asec:vis_res}) demonstrate a nearly comprehensive point cloud reconstruction, there are still some failure cases of Hestia (see~\cref{fig:fail}).
Hestia may occasionally fail to capture finer 3D structures, such as the window frames of the first-row house shown in~\cref{fig:fail}.
It may also fail to reconstruct highly self-occluded parts, such as the pillar of the second-row house.
In addition, Hestia sometimes struggles to capture bottom-up views that require extreme vertical viewing angles, for example, the Lego man's right hand and the underside of the pillar.
Moreover, it may struggle to reconstruct shapes with fine details over coarse surfaces, such as tiny parts of the broccoli and anise.
Adopting a multi-resolution grid structure or integrating octree-based methods to enhance the voxel grid resolution could be a potential future step to mitigate these issues.

In addition to the above limitations, we hope that Hestia will not be misused for other types of next-best-view (NBV) tasks.
In this study, we found that a close-greedy training scheme can effectively mitigate spurious correlations and is well-suited to our problem definition (see~\cref{asec:spur,sec:methods,sec:exp}).
However, the next-best-view problem is a broad research topic with varying objectives.
This finding may not generalize to other NBV tasks, such as next-best-view for object tracking or next-best-view for human aesthetics, where long-term planning is more critical.

Due to hardware limitations (e.g., the absence of an RGB-D camera), Hestia cannot fully exhibit its potential in the real-world drone system.
However, since we use a depth estimator to convert RGB images into depth maps, this limitation represents a trade-off rather than a fundamental constraint.
Our experiments conducted in NVIDIA IsaacLab demonstrate the full capability of Hestia, while the real-world application highlights Hestia's robustness when a depth sensor is unavailable.
Furthermore, due to drone regulations, the real-world application of Hestia is conducted indoors using an indoor GPS system (e.g., HTC Lighthouse base stations).
Drone policies vary across countries, and obtaining outdoor flight approvals can take up to a year in our region.
Additionally, outdoor trials require significant funding, such as renting a safe test site measuring approximately 100 meters by 100 meters.
As a future step, we plan to test Hestia outdoors to further validate its performance.
Another future step is to extend Hestia to a multi-agent setting for large-scale outdoor scanning (e.g., city-scale) under power-constrained scenarios.
